\documentclass[sigconf]{acmart}
\AtBeginDocument{%
  }

\setcopyright{acmlicensed}
\copyrightyear{2026}
\acmYear{2026}
\acmDOI{XXXXXXX.XXXXXXX}
\acmConference[Conference acronym 'XX]{Make sure to enter the correct
  conference title from your rights confirmation email}{June 03--05,
  2018}{Woodstock, NY}
\acmISBN{978-1-4503-XXXX-X/2018/06}

\usepackage{makecell}

\copyrightyear{2026}
\acmYear{2026}
\setcopyright{cc}
\setcctype{by-nc-nd}
\acmConference[WebSci Companion '26]{18th ACM Web Science Conference Companion}{May 26--29, 2026}{Braunschweig, Germany}
\acmBooktitle{18th ACM Web Science Conference Companion (WebSci Companion '26), May 26--29, 2026, Braunschweig, Germany}
\acmDOI{10.1145/3795513.3810448}
\acmISBN{979-8-4007-2492-3/2026/05}

\begin{document}


% NIcht finaler Titel
\title{Identifying Scientists on X}
% TItelvorschläge:
% Identifying Scientists on Twitter/X

%\acmSubmissionID 

% Reihenfolge der Autoren!!
\author{Philipp Meier}
\email{philipp.meier@hhu.de}
\affiliation{%
  \institution{Heinrich-Heine-University}
  \city{Düsseldorf}
  \country{Germany}
}

%\email{webmaster@marysville-ohio.com}
%\affiliation{%
%  \institution{Institute for Clarity in Documentation}
%  \country{USA}
%}

\author{Katarina Boland}
\email{katarina.boland@hhu.de}
\affiliation{%
  \institution{Heinrich-Heine-University}
  \city{Düsseldorf}
  \country{Germany}
}

\author{Laura Kallmeyer}
\email{kallmeyer@phil.uni-duesseldorf.de}
\affiliation{%
  \institution{Heinrich-Heine-University}
  \city{Düsseldorf}
  \country{Germany}
}

\author{Stefan Dietze}
\email{stefan.dietze@hhu.de}
\affiliation{
  \institution{Heinrich-Heine-University \& GESIS - Leibniz Institute for the Social Sciences, Cologne, Germany}
  \city{Düsseldorf}
  \country{Germany}
}


\begin{abstract}

%- Scope des Papers
%- Wir können zeigen dass, xy
%- Klassifikation der Usergruppen, keine tiefere Anaylse
% LLM Checked
With the growing importance of science-related discourse on the
Web and the erosion of the classical knowledge order, it is important
to identify different user groups, such as scientists, automatically.
This work proposes an approach for identifying scientists and non-
scientists on X/Twitter based on their user biographies and tweets.
We show that we are able to classify accounts as ’scientists’ and ’non-
scientists’ on two different datasets, reaching an F1 score of
up to 0.88 using Random Forests with linguistic features and up to
0.96 using a contrastively fine-tuned DeBERTa model in an ensemble setup. Furthermore,
we provide two datasets with X users labeled as scientists or non-
scientists and their respective tweets and user biographies.
\end{abstract}



%\ccsdesc[500]{Do Not Use This Code~Generate the Correct Terms for Your Paper}
%\ccsdesc[300]{Do Not Use This Code~Generate the Correct Terms for Your Paper}
%\ccsdesc{Do Not Use This Code~Generate the Correct Terms for Your Paper}
%\ccsdesc[100]{Do Not Use This Code~Generate the Correct Terms for Your Paper}


\received{20 February 2007}
\received[revised]{12 March 2009}
\received[accepted]{5 June 2009}

\maketitle

\section{Introduction}

%\begin{itemize}
%    \item How scientists use Twitter
%    \item Why is the automatic identification of scientists interesting?
%    \item Previous work: How did they approach the task?
%    \item Which gap do we fill with our contribution and what are our contribution
%\end{itemize}
%Within
% LLM checked
X, formerly Twitter, is a micro-blogging and social network platform. Scientists (among others) use such services to connect with other researchers, share and advertise papers or job opportunities. Especially in times of crisis like the COVID-19 pandemic, science communication on social networks got more and more important since public interest in science increased \citep{van2022reporting}. Scientists communicate more with laypeople, explaining scientific findings and research. At the same time, people with different backgrounds and occupations take part in science-related discourse. Through this rather informal discourse on the Web, scientific insights may be accidentally or deliberately oversimplified by or instrumentalized by different groups. With this paper, we present work in progress that aims at identifying scientists on X by using linguistic features as well as by using language models and contrastive learning. Automatically identifying scientists on social media allows to investigate patterns of %helps the understanding of
 scientific online discourse, and enables large scale automatic approaches. Specifically, it is a basis for understanding %Furthermore, it is important to understand
  scientists' interactions with other user groups, their roles, and aspects of %processing 
   trustworthiness and perceived expertise in general and science-related discourse. %automatically.

Especially during the COVID-19 pandemic, an increased public interest in scientific knowledge and science communication could be observed. %, which results in growing importance of science communication. 
Recent work by Biermann et al.\citep{biermann2023you, biermann2024does, biermann2025visible} investigated how scientists communicate in digital platforms in times of crisis. Furthermore, increasing public interest has also led to a dissolution of the traditional scientific knowledge order as recent work has shown \citep{fraser2021evolving,van2022reporting}, leading to oversimplification or politicization of scientific findings. This development is problematic since false information spreads faster than correct information as work of Vosoughi et al. \citep{vosoughi2018spread} shows. This indicates an urgency to identify and understand science-related online discourse.  % Hieraus lässt sich die Dringlichkeit Wissenschaftler automatisch zu erkennen, herleiten

%\noindent
% LLM checked
Previous work has focused on identifying scientists in a specific field, such as computer scientists \citep{hadgu2014identifying} or astrophysicists \citep{holmberg2014astrophysicists} or on a wide range of scientific fields, such as the natural, formal, and social sciences \citep{ke2017systematic}. Ke et al. \citep{ke2017systematic} found that fields such as mathematics or physics are under‑represented in online discourse, while social scientists are over‑represented.  
The online communication style of scientists is often investigated within communication science. While research in this area typically focuses on small samples of scientists and relies on manual annotation—as in Biermann et al. \citep{biermann2023you, biermann2024does, biermann2025visible}—there is other work that examines scientists tweeting during conferences \citep{bombaci2016using, allen2018twitter} or on a specific topic such as climate change \citep{walter2019scientific}.  
Moreover, with the growing interest in online science discourse, “perceived experts” are used by the public as information sources and can become vectors of misinformation, as shown in \citep{harris2024perceived}. This underscores why the identification of scientists is an especially important task in times of crisis.

%\noindent
However, to the best of our knowledge, there are no approaches yet which identify scientists without restrictions to academic fields or topics using merely information at the tweet and user level, i.e. not relying on network-based features or information such as Twitter/X Lists. We incorporate linguistic cues on tweet texts and user biographies to identify scientists without restrictions to topics or academic fields. %Previous work from Hadgu et al. \citep{hadgu2014identifying} focus on computer scientists and rely on profile features like the number of tweets or friends and content features, which include the usage of hashtags or retweets to tweets ratio among other features. 
 % by using linguistic cues from computational linguistics. 
We define ‘scientists’ as users who have an academic degree equal to or higher than a PhD and/or work as a scientist. We also consider persons who are currently doing a PhD as scientists. Accounts that belong to scientific entities, such as universities, research institutions or scientific conferences, also belong to the ‘scientist’ class. 
%Non-scientists are all other accounts.  Instances, where a classification as scientist or non-scientist is not possible, due to a lack of information, were labeled as 'Unknown'. These instances were merged with the non-scientists class.
We introduce a feature pipeline covering various types of lexical, syntactical, and semantical features to describe a user account on X, as well as several Pre-trained language model (PLM)-based models. Our best model using explicit features achieves an F1 score of 88\% for the identification of scientists and our DeBERTa Ensemble model an F1 score of 96\%. Besides our feature pipeline and our PLM-based classification models, we contribute two datasets named 'Orcid' and 'Scholar', annotated for the 'scientist' and 'non-scientist' classes, containing tweets and descriptions of users using TweetsKB \citep{fafalios2018tweetskb} as data source.

% Bitte noch nicht gegenlesen
\section{Related Work}
\label{sec:rel_work}

Recently, researchers in communication science focused on the communication style of scientists on social media. Biermann et al. \citep{biermann2023you, biermann2024does, biermann2025visible} focus on the communication style of scientists on X addressing climate change or political decisions during the COVID-19 pandemic. They worked in \citep{biermann2025visible} on a small set of scientists which were chosen manually. Furthermore, features like 'tonality' or 'uncertainty' were coded manually.\\ 
To expand such analysis to a larger scale, there exist also approaches that identify scientists automatically. However, such work often focused on scientists from a specific discipline such as in \citep{hadgu2014identifying}, where computer scientists are identified through matching candidate accounts with the DBLP database. These candidates were identified by collecting accounts that follow accounts which are typically followed by computer scientists. \citep{holmberg2014astrophysicists, haustein2014astrophysicists} focus on astrophysicists on X. In 2017, \citep{ke2017systematic} provide a large-scale approach covering scientists of scientific disciplines, covering natural, formal and social science. The authors used snowball sampling search on Twitter/X lists to identify scientists. Starting from seed users, breadth-first sampling over Twitter/X lists was performed. By using list memberships and list names containing scientist titles, they iteratively identified scientist accounts. %They found an over-representation of social scientists as well as an under representation of mathematical and physical scientists. 
In addition, the communication of scientists during conferences was covered by Allen et al. \citep{allen2018twitter} who analyzed scientists' tweets tweeted during a conference. Holmberg et al. \citep{holmberg2014astrophysicists} retrieved accounts of scientists in different research fields by using Web of Knowledge. Their work focused on the most productive scientists in each field. To identify such researchers, the count of publications in Web of Knowledge was used. However, this caused data sparsity since among the 20 most productive scientists there was only one scientist in astrophysics and none in economics. %In our approach, we did not want to focus on the most-productive researchers. 
Walter et al. \citep{walter2019scientific} focused on tweets by scientists about climate crisis. To identify scientists, the authors used information in the user biography. In our work, we also consider information in user biographies to identify scientists. Hashtags were also used to identify scientists as in a recent approach from Zhang et al. \citep{zhang2025online}, who identified potential scientists by first collecting accounts using specific hashtags to perform a manual coding. There exists also work about which user groups are reached by scientists, for example by \citep{bombaci2016using} who analyzed which user groups are reached by conference hashtags or by Cote et al. \citep{cote2018scientists} who analyzed which user groups are reached when scientists promote their work. Cote et al. found scientists by using 'Twitter/X lists', which are curated lists of users for a specific topic.\\

Previous approaches aiming to identify user groups on X also heavily relied on network information, like follower information or Twitter/X lists. For example, Wei et al. \citep{wei2016learning} exploit relations about followers of users and lists to identify topic experts on X. We cannot apply this approach because this information is not available anymore via the former Twitter Research API or in data archives such as TweetsKB. Cheng et al. \citep{cheng2014barbecue} also rely on Twitter/X lists to find local experts. Zeng et al. \citep{zeng2019detecting} identified journalists on social media by relying on self reported user information, network and Twitter/X-related features like the ratio of retweets. We use similar Twitter/X-related features like the count of hashtags or mentions. Closer to our approach is Khan et al. \citep{khan2016segregating}, who tried to identify bloggers and spammers using keywords. In our work, we also incorporate keywords for the identification of scientists. The problem of identifying scientists on social media is essentially an author-profiling task, where we want to infer the occupation of a user given textual input. PAN \citep{rangel2019overview} organized the shared task 'Author Profiling' from 2013 until 2020, where submissions had to predict attributes of authors like sex, age, occupation or if the author is a human or a bot using the provided data of PAN. Since 2015, competitors had to predict author attributes from tweets. Since information on networks or Twitter lists are not readily available anymore, we rely exclusively on tweet-level and user information to identify scientists. We approach the identification of scientists on X as an author profiling task, where the task is to predict whether an user is a scientist or not by using linguistic features on the tweet-level and keywords on the user biography as well as PLM-based embeddings.  \\

\section{Data}
%\begin{itemize}
%    \item We Build 2 datasets for researcher identification
    %\item Initial tries included keyword and emoji matches in a biography, which was later compared to a academic database. This caused a high false positive rate. 
%    \item First, this has lead to the creation of the Orcid dataset. Small dataset, where researchers are identified through their ORCID.
%    \item As non-scientist class for our ORCID dataset we used the previously defined 'non-scientists' class from %\citep{hadgu2014identifying}
%    \item Scholar, larger dataset, where researchers were identified by comparing their name to OpenAlex entries
%    \item Manual annotation of non-scientists sampled from TweetsKB to ensure data quality
%\end{itemize}
% LLM checked
In the following, we describe how we produced labeled data in order to train our scientist identification classifiers. To receive tweets from users by ID, we relied on TweetsKB \citep{hafid2022scitweets}, which is a corpus containing nearly 3.1 billion tweets spanning a time frame from January 2013 to June 2023. For each day in this time-frame, 1\% of all tweets were crawled and stored.
Identifying accounts belonging to scientists on X is not an easy task. We introduce two datasets for this purpose: Orcid and Scholar - building on previous publications of Hadgu et al. \cite{hadgu2014identifying}, and Mongeon et al. \cite{mongeon2023open}. %Since we are especially focusing on times of crisis, we are focusing
We focus on the time frame from December 2019 until May 2023. During this time of crisis (the COVID-19 pandemic), communication of scientists on social media gained prominence. 
%and recent findings like Biermann's \citep{biermann2023you, biermann2024does, biermann2025visible} were achieved in this time frame. 

%\begin{itemize}
%    \item Previous experiments with Keywords and Emojis to build a corpus
%    \item Ground truth missing, high false positive rate when comparing to scientific databases like OpenAlex or Google Scholar
%\end{itemize}

\subsection{Scientists}
\subsubsection{Orcid}
\noindent
% LLM checked
In order to find X accounts of scientists, we searched in TweetsKB for ORCID (Open Researcher and Contributor ID) matches in the account description, which ensures that the account belongs to a scientist. We used a regular expression to identify possible candidates and calculated the checksum of the candidate match to exclude false positive instances. A shortcoming of this matching method is that only a small number of scientists share their ORCID in the account description. Consequently, we missed many scientist accounts and, furthermore, our data is biased towards scientists with an ORCID specification in their account description. In total, this resulted in 578 matched accounts and 21,443 full text tweets.

%\begin{itemize}
%    \item What is an ORCID?
%    \item Used ORCID pattern used here % Link einfügen 
%    \item Bias: Only researchers who have an Orcid in their bio
%    \item Paired with data from \cite{hadgu2014identifying} % The jäschke data
%\end{itemize}
%LLM checked
\subsubsection{Scholar}
After building the Orcid corpus, we also built the Scholar corpus, which is based on the work of Mongeon et al. \cite{mongeon2023open}.
Mongeon et al. \cite{mongeon2023open} provide X account IDs of scientists in their corpus. As a first step, the authors searched for tweets where scientists announce or promote their publications by tweeting the DOI, using Crossref Event data dump from January 2022. This dump contained over 60 million Twitter/X events from 5,288,867 unique accounts, which contained the Tweet ID and the DOI of the paper mentioned in that tweet. Mongeon et al. assumed that among the users who tweet a publication, there will be at least one of the authors. To match the authors of the paper with the X account in order to identify the scientist, the authors searched the tweeted DOI in OpenAlex \citep{priem2022openalex} and compared the publication's author names to the X account name. OpenAlex is a database containing scholarly works assigned to authors and institutions. To do so, different matching steps were performed, for example by last name and full name, or by an exact match of the full name. This introduces a bias in this dataset, since it only contains scientists who tweeted about their publications by mentioning their DOIs. 
Since these matching methods do not provide a guarantee that the match between the X account and the OpenAlex entry is correct, Mongeon et al. performed a manual data validation on a subset of the matches with the three least precise matching steps between X account and OpenAlex entry. They verified that 61,645 instances were true positives.\\
%LLM checked
\noindent
We used the 61,645 true positive account IDs to find every tweet available in TweetsKB, and were able to hydrate 23,785 accounts. In order to prevent shortcut learning, we excluded tweets that contained a DOI from our dataset. This resulted in a total of 189,852 full text tweets. In terms of number of tweets per user, we observed a long-tail distribution, indicating that many accounts have a low count of tweets (e.g.~5 tweets), and only a few accounts have more than 100 tweets, as shown in Figure \ref{fig:res_stats}.%\\
% Ergänzung: DOIS exkludiert

\subsection{Non-Scientists}
\subsubsection{Orcid}
\label{sec:orcid}
%LLM checked
% Noch ergänzen wie die Researcher gefunden werden
% Auch hier im Set von Nicht-Wissenschaftlern könnten Wiss. enthalten sein, wurde diese überprüft?
Hadgu et al. \cite{hadgu2014identifying} identified computer scientists by matching candidate accounts with the database DBLP. To find non-scientist accounts, the authors crawled users via the Twitter/X Streaming API and removed every user who was contained in their set of scientists, thereby ensuring no overlap between scientists and non-scientists. In total, 1 million users were crawled and 1,500 accounts were sampled from this distribution as 'non-scientists'. It is not stated whether the 1,500 samples were manually annotated, so it could be the case that the negative samples could also contain accounts of scientists, especially from other disciplines than computer science. We hydrated these non-scientist accounts through TweetsKB and used the resulting accounts and tweets as the non-scientist class for our Orcid dataset. This resulted in 508 hydrated accounts and 6,737 tweets.

\subsubsection{Scholar}
\label{sec:scholar_samples}
\begin{figure}
    \centering
    \includegraphics[width=0.8\linewidth]{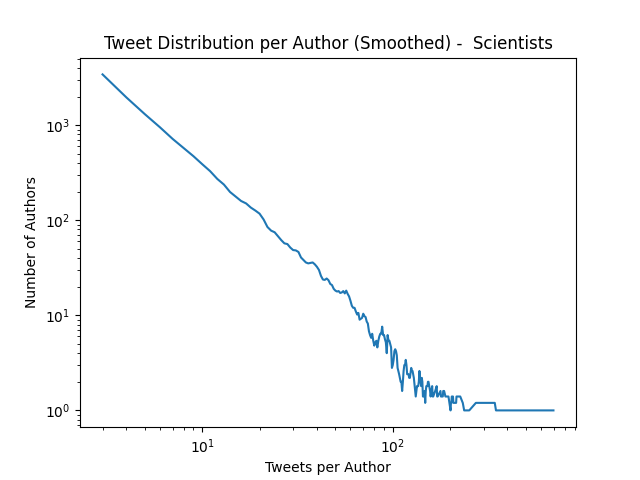}
    \caption{Log scaled distribution of tweets per user for scientists for the Scholar corpus}
    \label{fig:res_stats}
\end{figure}

\begin{figure}
    \centering
    \includegraphics[width=0.8\linewidth]{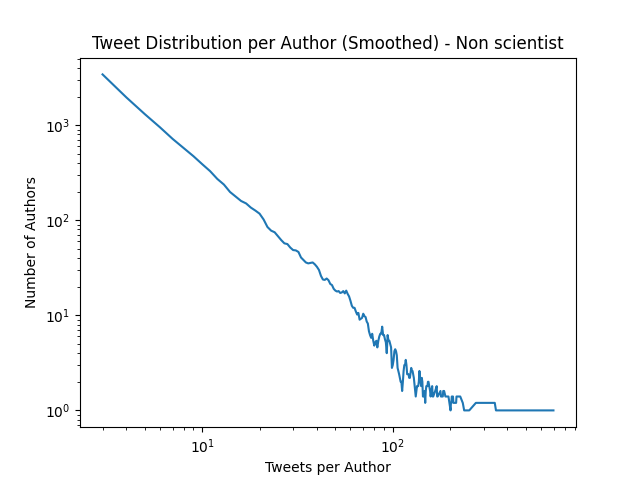}
    \caption{Log scaled distribution of tweets per user for non-scientists for the Scholar corpus}
    \label{fig:non_res_stats}
\end{figure}

% LLM checked
As the 'non-scientist' class, we took a random sample of user accounts from TweetsKB and rehydrated these accounts. We excluded users already present in the Orcid and Scholar scientist datasets. To ensure no imbalance between scientists and non-scientists regarding the number of tweets per user, the non-scientists data also follows a long-tail distribution. This was ensured by sampling the exact same number of X users with the exact same number of tweets as we have for scientists. The distribution is shown in Figure \ref{fig:res_stats} and Figure \ref{fig:non_res_stats}. It is important to note that we cannot fully be sure that we did not include scientists in this random sample, namely, scientists that are not included in our Scholar or ORCID scientist classes. 
% LLM checked
We conducted a manual annotation by two trained annotators (the first two authors of this paper) of a random sample of 400 users from the non-scientist class in the test set to estimate the number of scientists in our non-scientist class. Accounts were labeled as 'scientists', 'non-scientists' and 'unknown'. The label 'unknown' was used when there was no information indicative of the scientist class but there was no information which would rule out this possibility either, e.g. by detailing the user's occupation. After annotation, we added the instances of the unknown class to the non-scientist class. The annotators reached a kappa score of 0.67 and found 2 scientists in this sample. Furthermore, the same annotators also labeled 300 instances which were marked as 'false positive' by four different trained classifiers on the labeled test data (see section \ref{sec:res}, Table \ref{tab:schol_res}). We used the classifiers Random Forest, AdaBoost, Support Vector Machine and Logistic Regression on tweet-level. We found 10 scientists in this sample with an annotator agreement of 0.76. Note that this sample contains accounts with a high probability of being scientists and therefore serves as the estimate of an upper bound. Overall, this indicates that the quality of the non-scientist dataset is rather good. The annotation guidelines are available in our Github repository. 

%To quantify the amount of scientists in our non-scientists class, after training a classifier on the labeled data, we conducted a manual annotation by two trained annotators of 300 instances, which were marked as 'false positive' in our evaluation. 
% LLM checked
 In total, the dataset contains 379,704 tweets by 47,570 accounts. The dataset is balanced for scientists and non-scientists. To ensure a similar distribution as for the scientists partition, we counted the number of tweets per user and sampled the same number of users with the same tweet count for the non-scientists data. The distribution of users and their tweets counts between both classes is shown in Figure \ref{fig:res_stats} and \ref{fig:non_res_stats}. Both figures show a logarithmic scale and moving average smoothing was applied. %\\

%\noindent
% LLM checked
Statistics for the datasets are shown in Table \ref{tab:stats}. As shown, the Orcid dataset contains more than three times more tweets of scientists than tweets of non-scientists. The Scholar dataset is balanced in terms of amount of tweets and authors. For both datasets, we excluded retweets. 

\begin{table}[]
    \centering
    \begin{tabular}{|c|c|c|c|c|}
    \hline
         Data & \makecell{Tweets of \\scientists} & Scientists & \makecell{Tweets of \\non-scientists} & \makecell{Non-\\scientists}  \\ \hline
         Orcid & 21,443 &  578 & 6737 & 508  \\ % 28.180 508 non-scientists
         Scholar & 189,852 & 23,785 & 189,852 & 23,785 \\\hline
    \end{tabular}
    \caption{Dataset statistics for Orcid and Scholar datasets.}
    \label{tab:stats}
\end{table}

\subsubsection{Data examples}
% LLM checked
Table \ref{tab:data_examples} shows examples for the scientist, non-scientist and unknown classes. We choose only one tweet for each of the five different user accounts for demonstration purposes. However, for classification and manual annotations, all tweets of a user were considered.
The first example is clearly a scientist, which can be identified through cues in the biography (e.g. 'Professor'). In the second example, no direct information about the occupation is provided in the biography. However, the user's tweets indicate that the account probably belongs to a scientist. In the third case, the user provides information on their non-scientific occupation in the biography. In the fourth example, the tweets indicate that the account belongs to a student. In the selected tweet, the user asks for advice to get a good grade. %, because the person is not used to writing such papers. 
Examples three and four provide enough information to classify these users as non-scientists, since their occupation is either mentioned explicitly or we can infer it through their tweets. An 'unknown' case is displayed in the last row. Neither tweets nor author biography provide sufficient information to decide whether the person is a scientist or a non-scientist. 

%As shown, non‑scientist accounts can also post seemingly scientific tweets, such as the first tweet which is about the development of dataset sizes in machine learning for research. This tweet was labeled as 'scientific related' (class 3) by SciTweets. In the second tweet, a user quotes another tweet addressing vaccination. Both tweets are from our non-scientist dataset. The second tweet was annotated as 'scientific knowledge' by SciTweets. The remaining tweets are from scientists: the author of the third tweet provides advice to a colleague on how to improve the organization of student projects which may already appear as a scientist's tweet to the reader, while the last tweet could be from anyone as it addresses the global topic of engagement in online discourse. As the Table shows, the tweets from scientists were not annotated as 'scientific' by SciTweets.

\begin{table*}[]
    \scalebox{0.75}{
    \centering
    \begin{tabular}{|c|c|c|}
        \hline
        Tweet & Biography & Label  \\\hline
        %\makecell{REPLY: @user In my opinion, the free market will self-regulate in\\ this scenario, as the advancements  made by feeding \\large amounts of data into  neural networks will eventually reach a plateau. As a result,\\ research will shift towards smaller datasets and more comprehensible models.} & 3 & \makecell{Algorithm Development | Machine Learning | Mathematical Modeling \\ | Data Engineering | System Programming} & 0 \\\hline % Author: 1347754771860492289
        \makecell{Election Expert Richard L. Hasen on Voters’ Distrust [URL]} & \makecell{Professor of Law and Political Science at UC Irvine; \\ Election Law Blogger. Coming 2/4/20: \\ Election Meltdown (Preorders [URL])} & Scientist \\\hline

        %\makecell{You can’t stop being a #scientist just cuz you are on holiday!\\ Gorgeous #seaweed Codium fragile (otherwise known as green sea fingers,\\ dead man's fingers) at #sandycaperecreationpark near #jurienbaywa [URL] } & \makecell{Spatial is my speciality. If you are a \\coral, your location is a matter of life or death - especially\\ when a cyclone comes! Views my own. #TeamHB3} & Scientist \\\hline
        \makecell{\#AcademicTwitter - is there any advantage to editing a special\\ issue of a journal when you otherwise were going to publish your \\relevant work in other journals of much higher impact factor? } & \makecell{Spatial is my speciality. If you are a \\coral, your location is a matter of life or death - especially\\ when a cyclone comes! Views my own. \#TeamHB3} & Scientist \\\hline

        \makecell{@USER Here's just one of many articles that \\ debunk these stupid claims. Just Google it. [URL]} & \makecell{Software Engineer. Anything retweeted \\ is not adopted as a statement of fact.} & Non-Scientist\\\hline

        \makecell{Anyone used to writing qualitative research papers?\\ Writing this way is just not natural to me and \\ I'm struggling so if anyones familiar pls help me get an A} & keepin the fun in dysfunctional & Non-Scientist \\\hline

        REPLY: @USER @USER That’s my pick too,,, he would of been so perfect. & \makecell{Just a very cool Jim Morrison fan …..Cowboys Fan/ addicted to movies \& \\creating art, especially digital :: I Stand with Ukraine \& President Biden} & Unknown\\\hline
        %\makecell{QUOTE: Does this mean they will have herd immunity? Antibodies?\\ If so, would this be considered a silver lining?} & 1 & \makecell{Research, history, Photography, healthy cooking, weight training,\\ cycling, swimming, animals, debate, techie, science.\\ Devoted to my husband and to my freedom.} & 0 \\\hline % 40996431
        %\makecell{@user If your assessment is projects were comparatively lacking in quality,\\ then my recommendation would be to  think about how to \\ scaffold the project with milestones throughout the semester,\\ and consider ways to do so in cooperative or peer facilitated ways: [URL]} & 0 & \makecell{PhD Cand. and F31 fellow with @user @user studying \\ cancer evolution | Middling poet, reformed wrestler,\\ and pedagogist @user | @user alum} & 1\\\hline % 2571903062
        %\makecell{Its amazing how much stamina and energy you have when\\ its about someone wrong on the internet... (via @user) [URL] [URL]} & 0 & \makecell{Product exec; 8+ yrs as deeptech/x-border VC; working on \\ applying AI to medical data; lowkey-obsessed with \\ personal financeViews are mine; post/RT/ ≠ support} & 1  \\\hline % 15876871
    \end{tabular}}
    \caption{Example Tweets and user biographies from accounts with the label of the account. Users are anonymized with @USER, Urls with [URL].}
    \label{tab:data_examples}
\end{table*}

\section{Methods}
\label{sec:methods}

\subsection{Feature-based models}
\label{sec:methods}
%LLM checked
To identify scientists on X, we implemented a feature pipeline covering lexical, syntactic, X- and sentiment-based features, as well as topical features on tweet-level. A full list of features is provided in the Github repository \footnote{\url{https://github.com/PhMeier/identification_of_scientists_on_X.git}}. The features are calculated on the set of all tweets of a user. Biographies are often short, repetitive and free-from. We therefore decided not to apply our feature set for tweets on author biographies. Instead, to use author biography information in a meaningful way, we use keywords signaling a scientific profession and regex matches, described below. This allows a comparison between tweet-level and tweet-level with user descriptions.%To allow a comparison to pre-trained language models, which also use tweet and biography information and are described in section \ref{sec:plm}, we include the keyword and regex match feature. %As described in Related Work \label{sec:rel_work}, network features, which are often used in previous approaches, are not available in TweetsKB. Therefore, we use tweet-level information and user biography information in our proposed approach.%\\

%\noindent
Lexical features are common in tasks like authorship verification/attribution and author profiling. These features include (Corrected)-Type-Token Ratio (C)TTR), minimum and maximum word length, capitalization, spelling errors as well as readability metrics like Flesch-Kincaid or Adjusted Rand Index (ARI). To extract these features we used the lexical-richness library v.0.5.1 \citep{lex, accuracybias}, textstat v.0.7.3 \citep{textstat}, nltk v.3.8.1 \citep{loper2002nltk} and pyspellchecker v.0.8.2 \citep{spellchecker}. %\\

%\noindent
We used spaCy v.3.7.3 \citep{honnibal2020spacy} to extract syntactic features, including pronoun ratio and named entity ratio. Since Soler-Company et al. \citep{soler-company-wanner-2017-relevance} used complex syntactic features for author profiling and verification, we also included the depth of the dependency parse tree, the tree width (maximum number of nodes per level) and the ramification factor, which is the average number of children per level.%\\

%\noindent
X-based features capture the number of emojis, hashtags, mentions, and tweets in total. We also included two binary features for keyword and emoji pattern matches in user descriptions. Keywords are similar to those used in Harris et al.\citep{harris2024perceived}, where perceived experts were identified in scientific online discourse. As keywords we used "professor", "prof.", "prof", "scientist", "researcher", "postdoc", "dr.", "dr", "phd", "ph.d", "ph.d.". In addition, we search for emojis that may be related to science, such as a woman dressed as a scientist. These were manually selected by the first author. The keyword match and emoji pattern match features are binary, i.e. they capture whether at least one keyword or at least one science-related emoji is found in the biography, respectively. The complete list of keywords and emojis is available in the Github repository. %, so if one of the keywords is found, the keyword feature gets a value of one. Also, if one match or multiple matches of emojis are found, the emoji feature gets a value of one.

%\noindent
To compute sentiment-based features, we used the library \textit{tweetnlp} v.0.4.5 \citep{camacho-collados-etal-2022-tweetnlp}. These features capture on the one hand how emotional a text is, by capturing emotions like joy, anger, or anticipation. Additionally, these features also capture how vaguely a text is formulated, as well as whether a text is offensive or ironic. To capture how offensive a tweet is, we used the RoBERTa model \citep{liu2019roberta} by Barbieri et al. \citep{barbieri2020tweeteval}. To capture hedge we used the BERTweet model provided by Liew et al.\citep{hedgeBERT}.%\\

%\noindent
The topical features were created by using BERTopic v.0.16.3 \cite{grootendorst2022bertopic} with 50 topics. This model was trained on tweets of scientists from the Orcid dataset. We used standard parameters for the topic model with HDBSCAN \citep{mcinnes2017hdbscan} which allows a noise topic. We applied BERTopic to infer tweet-level topic assignments for each account. Tweets classified as outliers by HDBSCAN (topic -1) were excluded from the calculations. For all other tweets, we calculated absolute and relative topic frequencies on the account level. Additionally, we derived an account-level topic score by weighting each tweet’s topic probability with a topic weight based on the prevalence of that topic in the sample of the Orcid scientists, and then averaging the resulting values over tweets.
As additional topical feature, we used the 'science-relatedness' of tweets, determined by applying the SciTweets \cite{hafid2022scitweets} classifier. SciTweets is a BERT-based model finetuned on tweets annotated for science-relatedness. This covers four classes: Not-science related (class 0), Claim/Question (class 1), reference (class2) and research context (class 3). We computed how many tweets are classified as belonging to class 1, 2 and 3, respectively, and also computed the ratio by dividing the count by the total number of tweets.
%\noindent
As feature-based classification algorithms, we used the scikit-learn implementation \citep{scikit-learn} of Support Vector Machines (SVM), Random Forest, Logistic Regression, and AdaBoost. We chose SVM, Random Forest and Logistic Regression due to their strong performance in the author profiling shared task %Author Profiling
 at PAN \citep{rangel2019overview} \citep{hacohen2022survey}. We added AdaBoost as a classification algorithm because the setup includes two linear methods and two ensemble-based methods. % Motivation für Adaboost ergänzt

\subsection{Finetuned Pre-Trained Language Models}
\label{sec:plm}
In addition to the feature pipeline, we implemented two contrastively-fine-tuned language models.
We trained a DeBERTa model \citep{he2020deberta} with Contrastive Loss, and another model with Triplet Loss similar to \citep{rocca2022language}. We chose the DeBERTa model because it was a popular choice in PAN competitions for Writing Style tasks \citep{zangerle2024overview}. The models received representations of users as input, either the concatenated tweets of a user or their account descriptions. For each loss type, we implemented one model fine-tuned on tweets and one for user biographies. To make use of both tweet and biography information, we used both in an ensemble. As baseline, we used a plain DeBERTa model, which encoded tweets and author biographies in the same way as described in section \ref{sec:ensemble}. %The feature pipeline can also use biography information through the keyword and regex features.
We used the Sentence-Transformer library \citep{reimers-2019-sentence-bert} to train these models.%\\

\subsubsection{Contrastive Loss}
\label{sec:cl}
%\noindent
The goal of using the contrastive loss is that instances from the same class receive similar embeddings and instances from different classes obtain embeddings with an increased distance.  %should have different embeddings. 
More concretely, contrastive loss takes an input pair $(X_i, X_j)$ and minimizes the embedding distance when they are from the same class, while the distance between embeddings is increased if they are from different classes.
 
 %LLM checked
  The general form of Contrastive Loss \citep{hadsell2006dimensionality} is shown in Formula \ref{for:contrastive_loss}, where $(Y,X_1,X_2)^i$ describes the $i$-th input pair, consisting of the instances $X_1$, $X_2$ and the label $Y$. $P$ is the total number of input pairs. $D_W$ measures the cosine distance between $X_1$ and $X_2$, which are embedded by a parametric function $G$ which maps the input to a lower-dimensional space, where similar inputs should be close and dissimilar inputs are far apart: $G_W: R^D \rightarrow R^d$, where $d < D$. $W$ describes the shared parameters. $L_S$ is the partial loss function of similar pairs and $L_D$ the partial loss function for dissimilar pairs. The partial loss functions $L_S$ and $L_D$ are described through constraints in Hadsell et al: Both functions must be designed in such a way, that the minimization of $L$ decreases the distance between positive pairs and increase the distance of negative pairs. The definitions of both functions are shown below. $m$ in $L_D$ describes a margin, which must be greater than 0. The margin describes a radius around $G_W(X)$, where $X$ is an input vector. Two different partial loss functions are used because $L_D$, the loss function for the dissimilar pairs only contributes if its distance is within $m$, so in the radius of $G_W(X)$. According to the formula, similar pairs are labeled as '1' and dissimilar pairs are labeled as '0'.
  For pairs of the same class, we paired the representations of users and labeled them as a 'similar' pair. The representation of a user are either all concatenated tweets of a user or all concatenated user biographies of a user. For pairs of different classes (e.g., a pair of a scientist and non-scientist), we labeled these as 'dissimilar'.%\\

\begin{equation} 
\label{for:contrastive_loss}
\begin{split}
    %D_W(X_1, X_2) &= ||G_W(X_1) - G_W(X_2)||_2 \\ 
    D_W(X_1, X_2) &= 1 - cos(G_W(X_1),G_W(X_2))\\
    cos(G_W(X_1),G_W(X_2)) &= \frac{G_W(X_1) * G_W(X_2)}{||G_W(X_1)|| * ||G_W(X_2)||}\\
    L_S(W, X_1, X_2) &= \frac{1}{2}(D_W(X_1,X_2))^2\\
    L_D(W, X_1, X_2) &= \frac{1}{2}(max\{0,m-D_W(X_1,X_2)\})\\
    L(W) &= \sum^P_{i=1}L(W,(Y,X_1,X_2)^i) \\
    L(W,(Y,X_1,X_2)) &= Y*L_S(D_W) +\\
    &(1-Y) L_D(D_W)
\end{split}
\end{equation}

For inference, the contrastively fine-tuned model encodes the input $X$, where an instance is a user represented as concatenation of all of their tweets or user biography. These embeddings are fed then into a logistic regression model, which predicts 'scientist' or 'non-scientist'.
  
  %In contrastive learning, label '1' is used to decrease the distance between two embeddings and '0' is used to increase the distance between two embeddings. Transferred to our scenario, we want to increase the distance between embeddings of scientists and non-scientists, while decreasing the distance between embeddings of scientists and between non-scientists. Pairs of scientists and pairs of non-scientsts are labeled as 1, while pairs of scientists and non-scientists are labeled as 0. $L_G$ describes the partial loss for a positive pair, while $L_I$ describes the loss for a negative pair and $E_W$ is the similarity function between two pairs.

%\noindent
\subsubsection{Triplet Loss}
The triplet loss is also a contrastive learning function and was introduced in Schroff et al.\citep{schroff2015facenet}. The input is now a set of triplets $(a, p, n)$, where $a$, $p$ and $n$ are representations of three different users. Similar to the contrastive loss, we train two modality-specific encoders for triplet loss. More specifically, the triplets are formed from the concatenated texts of three distinct users, either all tweets or all biographies. 
$a$ describes an anchor, the user representation of a user %in our case concatenated tweets from a user
 of a class $j$. % or concatenated biographies from a user of a class $i$. 
  $p$ is a positive example, which is a user representation of a user %these are also concatenated tweets, but from another user
   which is also in class $j$. A negative example is given by $n$, which is a representation of a user that is not in class $j$. $N$ describes the size of the dataset and $i$ describes the instance of the dataset.
The triplet loss is shown in Formula \ref{for:triplet_loss}, where $X^a$ represents an anchor, $X^p$ a positive instance and $X^n$ a negative instance. $N$ is the size of the dataset, $||_2$ is the second euclidean norm, $\alpha$ describes a margin that is enforced between positive and negative pairs. Schroff et al. \citep{schroff2015facenet} set $\alpha$ to 0.2 in their work. The margin steers how much the negative example should be at least further away from the anchor than the positive instance. A small value can cause an overlap of positive and negative embeddings, which makes the embedding space unusable to distinguish between these classes. A high value can cause unstable training, since the model is forced to separate the positive and negative embeddings clearly, which possibly cannot be satisfied. $f(x)$ is a function representing the embedding. In our case, we used every tweet or every user description as a representation of a user. This representation of a user was then used to build triplets as described.%\\

\begin{equation} \label{for:triplet_loss}
\begin{split}
    \sum^N_i \left[||f(X^a_i) - f(X^p_i)||^2_2 - ||f(X^a_i)-f(X^n_i)||^2_2 + \alpha \right]
\end{split}
\end{equation}

For inference, apply the same procedure as described in section \ref{sec:cl}. Instead of a model fine-tuned by contrastive loss, we use the model trained with triplet loss.

%\noindent
We performed a hyperparameter search for the contrastively fine-tuned models. To find the optimal parameters, we used the first training fold split and divided this fold in 80\% training and 20\% validation. Best parameters for the contrastive learning model trained on bios (CL-BIO) had a learning rate of 2.93e-5, batch size of 16, gradient accumulation of 2, warmup ratio of 0.145, weight decay of 1.03e-06, linear learning rate scheduling, 5 training epochs and a margin of 0.64. The contrastive model trained on tweets used a learning rate of 4.48e-6
, batch size of 8, gradient accumulation of 2, warmup ratio of 0.07, weight decay of 1.5e-4, cosine learning rate scheduling, 3 training epochs and a margin of 0.42. The triplet model trained on biographies used a learning rate of 7.85e-6, a batch size of 8, 2 gradient accumulation steps, a warmup ratio of 0.10, a weight decay of 5.42e-05, cosine learning rate scheduling, 2 training epochs and a margin of 0.74. The triplet model trained on tweets used a learning rate of 2.05e-6, a batch size of 16, 4 gradient accumulation steps, a warmup ratio of 0.113, a weight decay of 5.71e-06, 3 training epochs and a margin of 0.46.
%\noindent 
\subsection{Ensemble of Language Models}
\label{sec:ensemble}
The contrastively fine-tuned models, which are either trained on tweets or on the user biographies, are combined through an ensemble classifier which uses the predictions of the models as input. Consider the following inputs: $X_{\mathit{tweets}}$, where each user is represented by a concatenation of their tweets and $X_{bio}$, where each user is represented by a concatenation of their biography texts. Since we are regarding the time frame from December 2019 until May 2023, a biography of a user may change over time. This means users can have multiple biography texts. $y_{tweets}$ or $y_{\mathit{bio}}$ are the labels of $X_{tweets}$ and $X_{bio}$. Each input type is fed into the respective contrastively trained model $CL_{tweets}$ or $CL_{bio}$, which produces embeddings $e_{tweets}$ or $e_{bio}$ for each instance in the dataset. These embeddings are used to train logistic regression models marked as $LR_{tweets}$ or $LR_{bio}$, which receive the corresponding embeddings $e$ and labels $y$. These logistic regression models output for each data instance a 2 dimensional array representing the probabilities for classes 'non-scientist' and 'scientist'. These outputs are $p_{tweets}$ for the logistic regression model receiving the embeddings of the contrastive model $CL_{tweets}$ trained on concatenated tweets and $p_{bio}$ for the model receiving the embeddings of $CL_{bio}$ as input.
These outputs are stacked in $z$, which is a 'meta-dataset' where each instance is represented as four-dimensional array containing the predictions of $LR_{tweets}$ and $LR_{bio}$. $z$ is used as input for the ensemble model $CL_{Ensemble}$ which is also a logistic regression model. $CL_{Ensemble}$ is trained on the predictions of $LR_{Tweets}$ and $LR_{Bio}$. Through this procedure, $CL_{Ensemble}$ combines the strength of each of our individual estimators.

\begin{equation}
    \begin{split}
    &e_{\mathit{tweets}} = CL_{\mathit{tweets}}(X_{\mathit{tweets}})\\
    &e_{\mathit{bio}} = CL_\mathit{bio}(X_{\mathit{bio}})\\
    &p_{\mathit{tweets}} = LR_\mathit{tweets}(e_\mathit{tweets}, y_\mathit{tweets})\\
    &p_{\mathit{bio}} = LR_\mathit{bio}(e_{\mathit{bio}}, y_{bio})\\
    &z = [p_{\mathit{tweets}}, p_{\mathit{bio}}]\\
    &CL_\mathit{Ensemble} = LR(z)\\
    \end{split}
    \label{for:stack}
\end{equation}

The ensemble can use predictions of the model trained on tweets and of the model trained on the user biography. Since the feature pipeline uses tweet information of users and can also use information from the biography through the features 'keywords' and 'emoji-match', we can compare our feature pipeline with the pre-trained language models.

%As baseline, a BERTweet model \citep{nguyen2020bertweet} was used. The same preprocessing of tweets and biographies was applied as described in Nguyen et al. \citep{nguyen2020bertweet}, which includes anonymization of users and Urls. The model received the concatenated tweets or bios of a user and had to classify the instance as scientists or non-scientist.

%\subsubsection{Evaluation Split}

%To ensure a comparison between the contrastive models to the feature pipeline The test set contains 3959 non-scientists and 4392 scientists.\\

%Therefore, we choose a logistic regression model because this model can return probabilities, which are used in our ensemble setup. 
% Noch Formel dazu gegeben

\section{Results} % Replaced
\label{sec:res}
\begin{table}[h!]
    \centering
    \begin{tabular}{|c|c|c|c|c|c}
    \hline
         Classifier & Accuracy & Precision & Recall & F1  \\\hline
         SVM & 0.84 & 0.85 &  0.82 & 0.84 \\
         Random Forest & 0.84 & 0.85 & 0.82 & 0.84 \\
         Ada Boost & 0.83 & 0.84 & 0.81 & 0.83 \\
         Logistic Reg. & 0.83 & 0.84 & 0.82 & 0.83 \\  \hline
         SVM + KW & 0.88 & 0.90 & 0.85 & 0.87 \\
         Random Forest + KW & \textbf{0.88} & \textbf{0.89} & \textbf{0.86} & \textbf{0.88} \\
         Ada Boost + KW & 0.88 & 0.89 & 0.86 & 0.87 \\
         Logistic Reg. + KW & 0.87 & 0.89 & 0.85 & 0.87 \\\hline
         only KW & 0.76 & 0.90 & 0.59 & 0.71 \\
         %SVM only KW & 0.68 & 0.75 & 0.67 & 0.65 \\
         %Random Forest only KW & 0.68 & 0.75 & 0.67 & 0.65 \\
         %Ada Boost only KW &  0.68 & 0.75 & 0.67 & 0.65 \\
         %Logistic Reg. only KW & 0.68 & 0.75 & 0.67 & 0.65  
         %\\ 
         \hline

    \end{tabular}
    \caption{Results on a 10-fold cross validation on the Scholar dataset. 'KW' stands for keyword features (Keyword+Regex), described in Section \ref{sec:methods}. Recall, Precision and F1 Scores are macro averaged.}
    \label{tab:schol_res}
\end{table}
% LLM checked
% replaced
The feature pipeline was evaluated in a 10-fold cross-validation setting on the Scholar and Orcid datasets. We compared 1) the classifiers using all features on the tweets of a user to 2) the classifiers using keywords and regex matches on the user biographies (marked as only KW) to 3) the classifiers using both (marked as + KW). Table \ref{tab:schol_res} shows the results on the Scholar dataset. All classifiers achieve similar results over the 10-fold cross validation. Random Forest achieves the highest results using the full feature set with an F1 score of 0.88. Without keywords as additional feature, Random Forest and SVM both yield the best result with an F1 score of 0.84. For all models using only the keyword and pattern match features, the same scores can be observed. We verified that this is due to them assigning every account with a match of either of these two features to the scientist class.\\
 %This indicates that solely relying on these features does not provide enough information for the models. Since the feature is binary, it does not offer enough discriminative characteristics for the instances.%\\

%\noindent
%LLM checked
% Replaced
For the Orcid dataset, the scores are lower overall. Again, all classifiers achieve similar results. For the complete feature set, AdaBoost achieves the best result with an F1 score of 0.82, while Random Forest achieves almost a similar result with an F1 score of 0.81. However, when not using keywords and regex pattern matches as features, the logistic regression achieves the best result with an F1 score of 0.79. Again, all models benefit from using keywords and regex pattern matches as additional features. For models using only the keywords as features, the same observation as for the Scholar dataset can be made. In combination with our feature set for tweets, the keyword features show the best performance.

% Replaced
\begin{table}[h]
    \centering
    \begin{tabular}{|c|c|c|c|c|c}
    \hline
         Classifier & Accuracy & Precision & Recall & F1  \\\hline
         SVM & 0.77 & 0.81 & 0.74 & 0.77 \\
         Random Forest & 0.77 & 0.79 & 0.78 & 0.78 \\
         Ada Boost & 0.76 & 0.78 & 0.77 & 0.77 \\
         Logistic Reg. & 0.78 & 0.80 & 0.79 & 0.79 \\ \hline
         SVM + KW & 0.81 & 0.84 & 0.79 & 0.81 \\
         Random Forest + KW & 0.80 & 0.83 & 0.79 & 0.81 \\
         Ada Boost + KW & \textbf{0.81} & \textbf{0.83} & \textbf{0.80} & \textbf{0.82} \\
         Logistic Reg. + KW & 0.81 & 0.83 & 0.80 & 0.81 \\ \hline
         %SVM only KW & 0.73 & 0.77 & 0.74 & 0.72 \\
         %Random Forest only KW & 0.73 & 0.77 & 0.74 & 0.72 \\
         %Ada Boost only KW & 0.73 & 0.77 & 0.74 & 0.72 \\
         %Logistic Reg. only KW & 0.73 & 0.77 & 0.74 & 0.72 \\
         only KW & 0.73 & 0.90 & 0.54 & 0.68 \\ \hline

    \end{tabular}
    \caption{Results on a 10-fold cross validation on the Orcid dataset. 'KW' stands for keyword features (Keyword+Regex), described in Section \ref{sec:methods}. Recall, Precision and F1 Scores are macro averaged.}
    \label{tab:orcid_res}
\end{table}

% Replaced
Table \ref{tab:cv_results_cl} shows the averaged results over 10 cross-validations on the Scholar dataset for the contrastive learners. The contrastively finetuned DeBERTa achieves the highest results as ensemble. The baseline and the triplet loss models achieve lower scores than the contrastive model. It requires further analysis why the triplet model achieves a lower score. The random forest model using keyword features achieves lower scores compared to the contrastive models in the ensemble setup with score of 0.84 F1 and 0.88, respectively. For both contrastive models, the scores are higher when user biography information is used instead of tweet information. This trend can be also observed for the DeBERTa baseline. This indicates that the biographies possibly encode more information about user identity than tweets. Furthermore, biographies tend to be relatively consistent over time, while the tweets of a user can cover various different topics. Combining predictions from tweet and biography models yields the highest results in the embedding model configuration. As shown in Table \ref{tab:data_examples}, tweets can contain complementary information, that allows a correct classification of users, which maybe misclassified if a model solely relies on biography information. While the contrastively finetuned DeBERTa model achieves the highest scores, pre-trained language models lack transparency. The Random Forest model, however, is a white box model that guarantees transparency by performing a feature analysis. Regarding previous work from  Hadgu et al. \citep{hadgu2014identifying} who focus on computer scientists and achieve an accuracy of 0.96 with their best model, we yield competitive results with an accuracy of 0.96 for scientists from unrestricted domains.

 % Replaced
\begin{table}[]
    \centering
    \begin{tabular}{|c|c|c|c|c|}
        \hline
         Model & Accuracy & Precision & Recall & F1  \\ \hline
         Random Forest & 0.84 & 0.85 & 0.82 & 0.84 \\
         Random Forest only KW & 0.76 & 0.90 & 0.57 & 0.71 \\
         Random Forest + KW & 0.88 & 0.89 & 0.86 & 0.88 \\\hline
         DeBERTa Ensemble & 0.94 & 0.94 & 0.95 & 0.94 \\  
         DeBERTa Tweet & 0.89 & 0.90 & 0.89 & 0.89\\ 
         DeBERTa Bio & 0.92 & 0.92 & 0.91 & 0.92\\ \hline
         CL Ensemble & \textbf{0.96} & \textbf{0.95} & \textbf{0.97} & \textbf{0.96}\\
         CL Tweets & 0.92 & 0.92 & 0.91 & 0.91 \\% 
         CL Bio & 0.95 & 0.96 & 0.93 & 0.95 \\\hline%  \hline
         Triplet Ensemble & 0.89 & 0.89 & 0.90 & 0.89 \\
         Triplet Tweets & 0.81 & 0.76 & 0.76 & 0.76 \\
         Triplet Bio & 0.83 & 0.82 & 0.86 & 0.84 \\\hline 
    \end{tabular} 
    \caption{Results of the contrastively fine-tuned models on the 10-fold cross validation on the Scholar dataset.  Precision, Recall and F1 are macro averaged. 'KW' stands for keyword features (Keyword+Regex).}
    \label{tab:cv_results_cl}
\end{table}        

%LLM checked

\begin{figure*}[h!]
    \centering
    \includegraphics[width=0.75\linewidth]{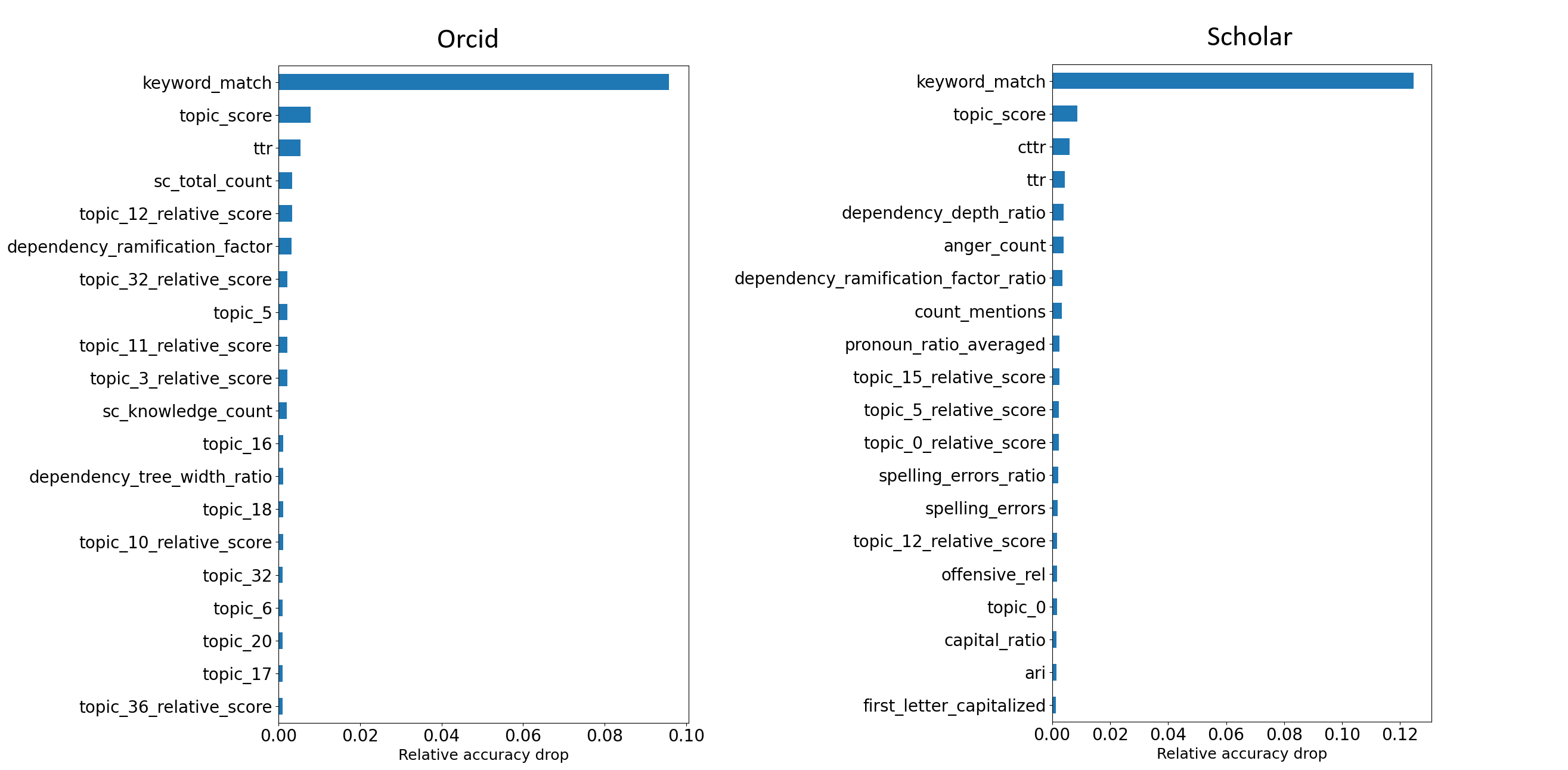}
    \caption{Top 10 most important features for Random Forest for the Orcid dataset and the Scholar dataset with keyword features}
    \label{fig:analysis_with_kw}
\end{figure*}

\section{Feature Analysis}
\subsection{Analysis with keyword features}
To see which linguistic features are actually important when classifying scientists, we analyzed the random forest model in the following way with feature permutation importance as implemented in scikit-learn \citep{scikit-learn}: We used the same 10-fold cross validation as used for the results in Tables \ref{tab:orcid_res} and \ref{tab:schol_res}. In each fold $k$, the feature values of feature $j$ were permuted. These scores $s_{k,j}$ were compared to the reference score $s$, which was computed using the non permuted feature. The difference between the reference score $s$ and the permuted score $s_{k,j}$ indicates the feature importance. Figure \ref{fig:analysis_with_kw} shows the mean feature importance over 10-fold cross validations for a random forest model with keyword features.%\\

%LLM checked
% Replaced
For both datasets, the most important feature is 'keyword match', which is not surprising, because the random forest classifier benefited from this feature on both datasets. 'Topic score' is also an important feature for both datasets. This feature captures how many tweets were assigned to one topic of the topic model described in section \ref{sec:methods}. The third feature of Scholar differs from Orcid. 'cttr' describes the corrected-type-token-ratio, while 'ttr' describes the type-token-ratio. Both features capture lexical diversity.'ttr' is considered as an important features on both dataset. By comparing both figures, one can see that complex syntactical features like dependency tree depth or the ramification factor play an important role in distinguishing scientists and non-scientists. However, for both figures, the relative accuracy drop is considerably lower for the displayed features compared to 'keyword match', which highlights the important role of keywords when identifying scientists.

\subsection{Analysis without keyword features}
\begin{figure*}[h!]
    \centering
    \includegraphics[width=0.75\linewidth]{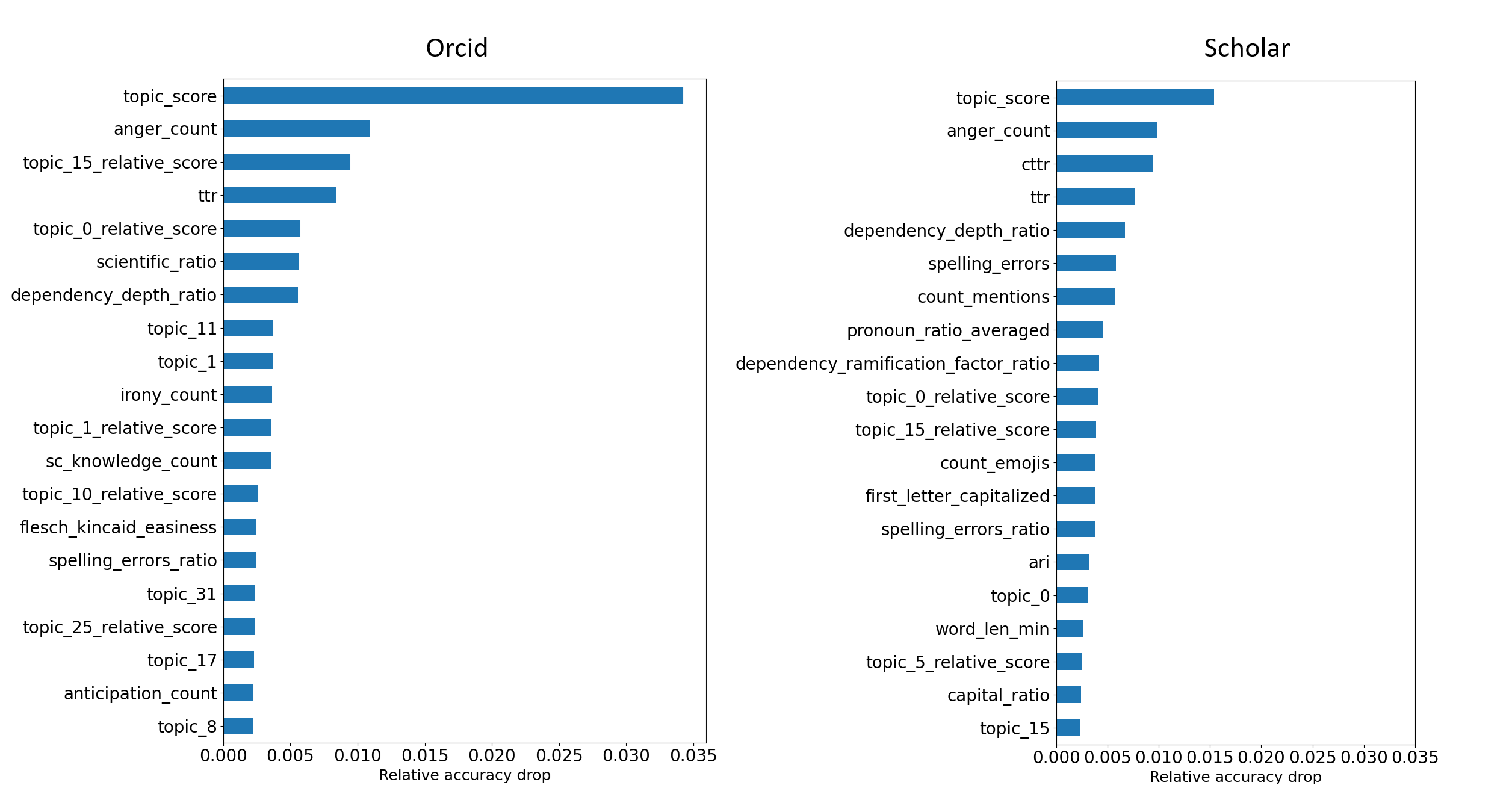}
    \caption{Top 10 most important features for Random Forest for the Orcid dataset and the Scholar dataset without keyword features}
    \label{fig:analysis_without_kw}
\end{figure*}

% Repalced
Excluding the keyword features 'keyword-matches' and 'regex-matches' leads to a different distribution of features on both datasets. While the four most important features 'CTTR', 'topic score', 'anger count', and 'ttr', are similar to those on the Scholar dataset when using keywords and regex patterns as features, one can see that topic related features play a more significant role when not using keywords, as Figure \ref{fig:analysis_without_kw} shows. The Random Forest model also seems to rely more on lexical features when keywords are not included, regarding the higher scores of 'spelling errors' and 'first letter capitalized' with with scores of 0.005. When using keyword features, both features had a low score of nearly zero. For the Orcid dataset, the reliance on lexical features is even more concise: 'TTR' has a value of around 0.007 and is the fourth most important feature and the Flesch-Kincaid Reading Easiness score has a score around 0.0025. The average pronoun ratio also plays a significant role on the scholar dataset. It can also be observed that sentiment based features like 'irony count' or 'anticipation count', which represent the number of  ironic or tweets which anticipate something, are more important on the Orcid dataset, than on the Scholar dataset. A striking difference is the significance of 'scientific ratio', the amount of tweets that were labeled as 'scientific' by the SciTweets model: This feature is among the top 10 features on the Orcid dataset, but not on the Scholar dataset.

% Repalced
Biermann et al. \citep{biermann2025visible} found that scientists pay attention to correct spelling. The importance of the features 'spelling errors' and 'first letter capitalized' is in line with this. Moreover, they found that scientists rarely communicated scientific uncertainty in their tweets. We used the features 'hedge count' and 'hedge rel' as a proxy for this communication style. As Figure \ref{fig:analysis_with_kw} and \ref{fig:analysis_without_kw} show, these features are not present in the top 10 features. Thus, they do not seem to discriminate well between the two classes. We further hypothesized that scientists tend to talk more about specific, potentially science-related topics. This hypothesis is supported by the importance of topical features. 

\section{Discussion}
%LLM checked
%LLM checked
% Replaced
The results show how powerful linguistic features are on tweet-level when identifying scientists on X, even without using keywords as an additional feature. While the contrastively finetuned DeBERTa model yields the highest results, such approaches lack transparency. Since we are also interested in which linguistic features are reliable and significant when classifying scientists, further analysis will focus on our implemented feature pipeline, e.g. applying our pipeline to other user groups to investigate to which extent certain features correlate with specific groups. As mentioned in Section 'Data' \ref{sec:scholar_samples}, the randomly sampled accounts, which are used as non-scientists, possibly include scientists. To quantify the amount of those scientists, we performed a manual annotation on 400 randomly chosen samples from the Scholar dataset. The results indicate that the evaluation results should not be biased substantially by false negative instances. Still, a bias cannot be ruled out. The negative samples used in the Orcid dataset may also contain accounts of scientists, as mentioned in section \ref{sec:orcid}. %As mentioned in section \ref{sec:scholar_samples}, our sampled set from TweetsKB contains researcher. However, from 400 random samples from TweetsKB, only two accounts were considered as researcher. 
Our approach can only detect scientists who share information about their profession. Scientists who do not directly share such information in their biographies or tweet about their research or employment cannot be identified. Furthermore, the applicability to other platforms like Mastodon is currently unknown.
Considering the ethical implications of our work, we will restrict access to the dataset and will not publicly share UserIDs or personal information associated with the accounts to prevent users from becoming targets of repression or harassment.

\section{Future Work}
%LLM checked
In future work, we plan to increase the quality of our dataset by adding additional annotations. Furthermore, we want to extend our classification to two new classes, namely journalists and politicians. %We will build our dataset of journalists on work of \citep{toprak2022journalists} and for politicians on work of \citep{bor2023quantifying}. 
Regarding classification results, we want to further analyze which features are strong and weak when classifying different user groups and further improve our contrastively fine-tuned models, for example by different data augmentation methods. Finally, we want to combine our feature-based approach with our PLM-based models in an ensemble to build a strong classifier.

\section{Conclusion}
%LLM checked
In this paper, we have shown how scientists can be detected by using linguistic cues in tweet texts. We could see that additional features like keyword and regex matches on the user biographies, may, but do not always, lead to an improvement of models. The feature analysis revealed that lexical features play a substantial role for both of our datasets when classifying scientists. Furthermore, we could observe the potential of contrastive learning in this setting. However, models using implicit features yield better results than models using explicit features. We plan to further investigate which features are helpful in the identification of scientists and to apply our approach to other user groups like journalists or politicians.

\section*{Acknowledgments}
 
We would like to thank the anonymous reviewers for their useful feedback. This work was carried out in the project \textit{NewOrder -- Understanding the erosion of the traditional knowledge order in scientific online discourse and its impact in times of crisis} (project number: K490/2022) funded by the Leibniz Association.


%\begin{acks}
%This work was carried out in the project \textit{NewOrder -- Understanding the erosion of the traditional knowledge order in scientific online discourse and its impact in times of crisis} (project number: K490/2022) funded by the Leibniz Association.
%\end{acks}

\bibliographystyle{ACM-Reference-Format}
\bibliography{my_bib}


\begin{thebibliography}{47}

%%% the Web address.

\ifx \showCODEN    \undefined \def \showCODEN     #1{\unskip}     \fi
\ifx \showISBNx    \undefined \def \showISBNx     #1{\unskip}     \fi
\ifx \showISBNxiii \undefined \def \showISBNxiii  #1{\unskip}     \fi
\ifx \showISSN     \undefined \def \showISSN      #1{\unskip}     \fi
\ifx \showLCCN     \undefined \def \showLCCN      #1{\unskip}     \fi
\ifx \shownote     \undefined \def \shownote      #1{#1}          \fi
\ifx \showarticletitle \undefined \def \showarticletitle #1{#1}   \fi
\ifx \showURL      \undefined \def \showURL       {\relax}        \fi
% invisible to TeX
\providecommand\bibfield[2]{#2}
\providecommand\bibinfo[2]{#2}
\providecommand\natexlab[1]{#1}
\providecommand\showeprint[2][]{arXiv:#2}

\bibitem[Allen et~al\mbox{.}(2018)]%
        {allen2018twitter}
\bibfield{author}{\bibinfo{person}{Caitlin~G Allen}, \bibinfo{person}{Brittany Andersen}, \bibinfo{person}{David~A Chambers}, \bibinfo{person}{Jacob Groshek}, {and} \bibinfo{person}{Megan~C Roberts}.} \bibinfo{year}{2018}\natexlab{}.
\newblock \showarticletitle{Twitter use at the 2016 Conference on the Science of Dissemination and Implementation in Health: analyzing\# DIScience16}.
\newblock \bibinfo{journal}{\emph{Implementation Science}} \bibinfo{volume}{13}, \bibinfo{number}{1} (\bibinfo{year}{2018}), \bibinfo{pages}{34}.
\newblock


\bibitem[Bansal and Aggarwal(2024)]%
        {textstat}
\bibfield{author}{\bibinfo{person}{Shivam Bansal} {and} \bibinfo{person}{Chaitanya Aggarwal}.} \bibinfo{year}{2024}\natexlab{}.
\newblock \bibinfo{title}{Textstat}.
\newblock
\urldef\tempurl%
\url{https://textstat.org/}
\showURL{%
\tempurl}


\bibitem[Barbieri et~al\mbox{.}(2020)]%
        {barbieri2020tweeteval}
\bibfield{author}{\bibinfo{person}{Francesco Barbieri}, \bibinfo{person}{Jose Camacho-Collados}, \bibinfo{person}{Luis~Espinosa Anke}, {and} \bibinfo{person}{Leonardo Neves}.} \bibinfo{year}{2020}\natexlab{}.
\newblock \showarticletitle{TweetEval: Unified benchmark and comparative evaluation for tweet classification}. In \bibinfo{booktitle}{\emph{Findings of the association for computational linguistics: EMNLP 2020}}. \bibinfo{pages}{1644--1650}.
\newblock


\bibitem[Barrus(2018)]%
        {spellchecker}
\bibfield{author}{\bibinfo{person}{Tyler Barrus}.} \bibinfo{year}{2018}\natexlab{}.
\newblock \bibinfo{title}{Spellchecker}.
\newblock
\urldef\tempurl%
\url{https://pypi.org/project/pyspellchecker/}
\showURL{%
\tempurl}


\bibitem[Biermann et~al\mbox{.}(2024)]%
        {biermann2024does}
\bibfield{author}{\bibinfo{person}{Kaija Biermann}, \bibinfo{person}{Bianca Nowak}, \bibinfo{person}{Lea-Marie Braun}, \bibinfo{person}{Monika Taddicken}, \bibinfo{person}{Nicole~C Kr{\"a}mer}, {and} \bibinfo{person}{Stefan Stieglitz}.} \bibinfo{year}{2024}\natexlab{}.
\newblock \showarticletitle{Does scientific evidence sell? Combining manual and automated content analysis to investigate scientists’ and laypeople’s evidence practices on social media}.
\newblock \bibinfo{journal}{\emph{Science Communication}} \bibinfo{volume}{46}, \bibinfo{number}{5} (\bibinfo{year}{2024}), \bibinfo{pages}{619--652}.
\newblock


\bibitem[Biermann et~al\mbox{.}(2023)]%
        {biermann2023you}
\bibfield{author}{\bibinfo{person}{Kaija Biermann}, \bibinfo{person}{Nicola Peters}, {and} \bibinfo{person}{Monika Taddicken}.} \bibinfo{year}{2023}\natexlab{}.
\newblock \showarticletitle{“You Can Do Better Than That!”: Tweeting Scientists Addressing Politics on Climate Change and Covid-19}.
\newblock \bibinfo{journal}{\emph{Media and Communication}} \bibinfo{volume}{11}, \bibinfo{number}{1} (\bibinfo{year}{2023}), \bibinfo{pages}{217--227}.
\newblock


\bibitem[Biermann and Taddicken(2025)]%
        {biermann2025visible}
\bibfield{author}{\bibinfo{person}{Kaija Biermann} {and} \bibinfo{person}{Monika Taddicken}.} \bibinfo{year}{2025}\natexlab{}.
\newblock \showarticletitle{Visible scientists in digital communication environments: An analysis of their role performance as public experts on Twitter/X during the Covid-19 pandemic}.
\newblock \bibinfo{journal}{\emph{Public Understanding of Science}} \bibinfo{volume}{34}, \bibinfo{number}{1} (\bibinfo{year}{2025}), \bibinfo{pages}{38--58}.
\newblock


\bibitem[Bombaci et~al\mbox{.}(2016)]%
        {bombaci2016using}
\bibfield{author}{\bibinfo{person}{Sara~P Bombaci}, \bibinfo{person}{Cooper~M Farr}, \bibinfo{person}{H~Travis Gallo}, \bibinfo{person}{Anna~M Mangan}, \bibinfo{person}{Lani~T Stinson}, \bibinfo{person}{Monica Kaushik}, {and} \bibinfo{person}{Liba Pejchar}.} \bibinfo{year}{2016}\natexlab{}.
\newblock \showarticletitle{Using Twitter to communicate conservation science from a professional conference}.
\newblock \bibinfo{journal}{\emph{Conservation Biology}} \bibinfo{volume}{30}, \bibinfo{number}{1} (\bibinfo{year}{2016}), \bibinfo{pages}{216--225}.
\newblock


\bibitem[Camacho-collados et~al\mbox{.}(2022)]%
        {camacho-collados-etal-2022-tweetnlp}
\bibfield{author}{\bibinfo{person}{Jose Camacho-collados}, \bibinfo{person}{Kiamehr Rezaee}, \bibinfo{person}{Talayeh Riahi}, \bibinfo{person}{Asahi Ushio}, \bibinfo{person}{Daniel Loureiro}, \bibinfo{person}{Dimosthenis Antypas}, \bibinfo{person}{Joanne Boisson}, \bibinfo{person}{Luis Espinosa~Anke}, \bibinfo{person}{Fangyu Liu}, {and} \bibinfo{person}{Eugenio Mart{\'i}nez~C{\'a}mara}.} \bibinfo{year}{2022}\natexlab{}.
\newblock \showarticletitle{{T}weet{NLP}: Cutting-Edge Natural Language Processing for Social Media}. In \bibinfo{booktitle}{\emph{Proceedings of the 2022 Conference on Empirical Methods in Natural Language Processing: System Demonstrations}}, \bibfield{editor}{\bibinfo{person}{Wanxiang Che} {and} \bibinfo{person}{Ekaterina Shutova}} (Eds.). \bibinfo{publisher}{Association for Computational Linguistics}, \bibinfo{address}{Abu Dhabi, UAE}, \bibinfo{pages}{38--49}.
\newblock
\href{https://doi.org/10.18653/v1/2022.emnlp-demos.5}{doi:\nolinkurl{10.18653/v1/2022.emnlp-demos.5}}


\bibitem[Cheng et~al\mbox{.}(2014)]%
        {cheng2014barbecue}
\bibfield{author}{\bibinfo{person}{Zhiyuan Cheng}, \bibinfo{person}{James Caverlee}, \bibinfo{person}{Himanshu Barthwal}, {and} \bibinfo{person}{Vandana Bachani}.} \bibinfo{year}{2014}\natexlab{}.
\newblock \showarticletitle{Who is the barbecue king of texas? A geo-spatial approach to finding local experts on twitter}. In \bibinfo{booktitle}{\emph{Proceedings of the 37th international ACM SIGIR conference on Research \& development in information retrieval}}. \bibinfo{pages}{335--344}.
\newblock


\bibitem[C{\^o}t{\'e} and Darling(2018)]%
        {cote2018scientists}
\bibfield{author}{\bibinfo{person}{Isabelle~M C{\^o}t{\'e}} {and} \bibinfo{person}{Emily~S Darling}.} \bibinfo{year}{2018}\natexlab{}.
\newblock \showarticletitle{Scientists on Twitter: Preaching to the choir or singing from the rooftops?}
\newblock \bibinfo{journal}{\emph{Facets}} \bibinfo{volume}{3}, \bibinfo{number}{1} (\bibinfo{year}{2018}), \bibinfo{pages}{682--694}.
\newblock


\bibitem[Fafalios et~al\mbox{.}(2018)]%
        {fafalios2018tweetskb}
\bibfield{author}{\bibinfo{person}{Pavlos Fafalios}, \bibinfo{person}{Vasileios Iosifidis}, \bibinfo{person}{Eirini Ntoutsi}, {and} \bibinfo{person}{Stefan Dietze}.} \bibinfo{year}{2018}\natexlab{}.
\newblock \showarticletitle{Tweetskb: A public and large-scale rdf corpus of annotated tweets}. In \bibinfo{booktitle}{\emph{European Semantic Web Conference}}. Springer, \bibinfo{pages}{177--190}.
\newblock


\bibitem[Fraser et~al\mbox{.}(2021)]%
        {fraser2021evolving}
\bibfield{author}{\bibinfo{person}{Nicholas Fraser}, \bibinfo{person}{Liam Brierley}, \bibinfo{person}{Gautam Dey}, \bibinfo{person}{Jessica~K Polka}, \bibinfo{person}{M{\'a}t{\'e} P{\'a}lfy}, \bibinfo{person}{Federico Nanni}, {and} \bibinfo{person}{Jonathon~Alexis Coates}.} \bibinfo{year}{2021}\natexlab{}.
\newblock \showarticletitle{The evolving role of preprints in the dissemination of COVID-19 research and their impact on the science communication landscape}.
\newblock \bibinfo{journal}{\emph{PLoS biology}} \bibinfo{volume}{19}, \bibinfo{number}{4} (\bibinfo{year}{2021}), \bibinfo{pages}{e3000959}.
\newblock


\bibitem[Grootendorst(2022)]%
        {grootendorst2022bertopic}
\bibfield{author}{\bibinfo{person}{Maarten Grootendorst}.} \bibinfo{year}{2022}\natexlab{}.
\newblock \showarticletitle{BERTopic: Neural topic modeling with a class-based TF-IDF procedure}.
\newblock \bibinfo{journal}{\emph{arXiv preprint arXiv:2203.05794}} (\bibinfo{year}{2022}).
\newblock


\bibitem[HaCohen-Kerner(2022)]%
        {hacohen2022survey}
\bibfield{author}{\bibinfo{person}{Yaakov HaCohen-Kerner}.} \bibinfo{year}{2022}\natexlab{}.
\newblock \showarticletitle{Survey on profiling age and gender of text authors}.
\newblock \bibinfo{journal}{\emph{Expert Systems with Applications}}  \bibinfo{volume}{199} (\bibinfo{year}{2022}), \bibinfo{pages}{117140}.
\newblock


\bibitem[Hadgu and J{\"a}schke(2014)]%
        {hadgu2014identifying}
\bibfield{author}{\bibinfo{person}{Asmelash~Teka Hadgu} {and} \bibinfo{person}{Robert J{\"a}schke}.} \bibinfo{year}{2014}\natexlab{}.
\newblock \showarticletitle{Identifying and analyzing researchers on twitter}. In \bibinfo{booktitle}{\emph{Proceedings of the 2014 ACM conference on Web science}}. \bibinfo{pages}{23--32}.
\newblock


\bibitem[Hadsell et~al\mbox{.}(2006)]%
        {hadsell2006dimensionality}
\bibfield{author}{\bibinfo{person}{Raia Hadsell}, \bibinfo{person}{Sumit Chopra}, {and} \bibinfo{person}{Yann LeCun}.} \bibinfo{year}{2006}\natexlab{}.
\newblock \showarticletitle{Dimensionality reduction by learning an invariant mapping}. In \bibinfo{booktitle}{\emph{2006 IEEE computer society conference on computer vision and pattern recognition (CVPR'06)}}, Vol.~\bibinfo{volume}{2}. IEEE, \bibinfo{pages}{1735--1742}.
\newblock


\bibitem[Hafid et~al\mbox{.}(2022)]%
        {hafid2022scitweets}
\bibfield{author}{\bibinfo{person}{Salim Hafid}, \bibinfo{person}{Sebastian Schellhammer}, \bibinfo{person}{Sandra Bringay}, \bibinfo{person}{Konstantin Todorov}, {and} \bibinfo{person}{Stefan Dietze}.} \bibinfo{year}{2022}\natexlab{}.
\newblock \showarticletitle{Scitweets-a dataset and annotation framework for detecting scientific online discourse}. In \bibinfo{booktitle}{\emph{Proceedings of the 31st ACM International Conference on Information \& Knowledge Management}}. \bibinfo{pages}{3988--3992}.
\newblock


\bibitem[Harris et~al\mbox{.}(2024)]%
        {harris2024perceived}
\bibfield{author}{\bibinfo{person}{Mallory~J Harris}, \bibinfo{person}{Ryan Murtfeldt}, \bibinfo{person}{Shufan Wang}, \bibinfo{person}{Erin~A Mordecai}, {and} \bibinfo{person}{Jevin~D West}.} \bibinfo{year}{2024}\natexlab{}.
\newblock \showarticletitle{Perceived experts are prevalent and influential within an antivaccine community on Twitter}.
\newblock \bibinfo{journal}{\emph{PNAS nexus}} \bibinfo{volume}{3}, \bibinfo{number}{2} (\bibinfo{year}{2024}), \bibinfo{pages}{pgae007}.
\newblock


\bibitem[Haustein et~al\mbox{.}(2014)]%
        {haustein2014astrophysicists}
\bibfield{author}{\bibinfo{person}{Stefanie Haustein}, \bibinfo{person}{Timothy~D Bowman}, \bibinfo{person}{Kim Holmberg}, \bibinfo{person}{Isabella Peters}, {and} \bibinfo{person}{Vincent Larivi{\`e}re}.} \bibinfo{year}{2014}\natexlab{}.
\newblock \showarticletitle{Astrophysicists on Twitter: An in-depth analysis of tweeting and scientific publication behavior}.
\newblock \bibinfo{journal}{\emph{Aslib Journal of Information Management}} \bibinfo{volume}{66}, \bibinfo{number}{3} (\bibinfo{year}{2014}), \bibinfo{pages}{279--296}.
\newblock


\bibitem[He et~al\mbox{.}(2020)]%
        {he2020deberta}
\bibfield{author}{\bibinfo{person}{Pengcheng He}, \bibinfo{person}{Xiaodong Liu}, \bibinfo{person}{Jianfeng Gao}, {and} \bibinfo{person}{Weizhu Chen}.} \bibinfo{year}{2020}\natexlab{}.
\newblock \showarticletitle{Deberta: Decoding-enhanced bert with disentangled attention}.
\newblock \bibinfo{journal}{\emph{arXiv preprint arXiv:2006.03654}} (\bibinfo{year}{2020}).
\newblock


\bibitem[Holmberg et~al\mbox{.}(2014)]%
        {holmberg2014astrophysicists}
\bibfield{author}{\bibinfo{person}{Kim Holmberg}, \bibinfo{person}{Timothy~D Bowman}, \bibinfo{person}{Stefanie Haustein}, {and} \bibinfo{person}{Isabella Peters}.} \bibinfo{year}{2014}\natexlab{}.
\newblock \showarticletitle{Astrophysicists’ conversational connections on Twitter}.
\newblock \bibinfo{journal}{\emph{PloS one}} \bibinfo{volume}{9}, \bibinfo{number}{8} (\bibinfo{year}{2014}), \bibinfo{pages}{e106086}.
\newblock


\bibitem[Honnibal et~al\mbox{.}(2020)]%
        {honnibal2020spacy}
\bibfield{author}{\bibinfo{person}{Matthew Honnibal}, \bibinfo{person}{Ines Montani}, \bibinfo{person}{Sofie Van~Landeghem}, {and} \bibinfo{person}{Adriane Boyd}.} \bibinfo{year}{2020}\natexlab{}.
\newblock \showarticletitle{{spaCy: Industrial-strength Natural Language Processing in Python}}.
\newblock  (\bibinfo{year}{2020}).
\newblock
\href{https://doi.org/10.5281/zenodo.1212303}{doi:\nolinkurl{10.5281/zenodo.1212303}}


\bibitem[Ke et~al\mbox{.}(2017)]%
        {ke2017systematic}
\bibfield{author}{\bibinfo{person}{Qing Ke}, \bibinfo{person}{Yong-Yeol Ahn}, {and} \bibinfo{person}{Cassidy~R Sugimoto}.} \bibinfo{year}{2017}\natexlab{}.
\newblock \showarticletitle{A systematic identification and analysis of scientists on Twitter}.
\newblock \bibinfo{journal}{\emph{PLoS one}} \bibinfo{volume}{12}, \bibinfo{number}{4} (\bibinfo{year}{2017}), \bibinfo{pages}{e0175368}.
\newblock


\bibitem[Khan et~al\mbox{.}(2016)]%
        {khan2016segregating}
\bibfield{author}{\bibinfo{person}{Muhammad Usman~Shahid Khan}, \bibinfo{person}{Mazhar Ali}, \bibinfo{person}{Assad Abbas}, \bibinfo{person}{Samee~U Khan}, {and} \bibinfo{person}{Albert~Y Zomaya}.} \bibinfo{year}{2016}\natexlab{}.
\newblock \showarticletitle{Segregating spammers and unsolicited bloggers from genuine experts on twitter}.
\newblock \bibinfo{journal}{\emph{IEEE Transactions on Dependable and Secure Computing}} \bibinfo{volume}{15}, \bibinfo{number}{4} (\bibinfo{year}{2016}), \bibinfo{pages}{551--560}.
\newblock


\bibitem[Liew({[n.\,d.]})]%
        {hedgeBERT}
\bibfield{author}{\bibinfo{person}{Christopher Liew}.} \bibinfo{year}{[n.\,d.]}\natexlab{}.
\newblock \bibinfo{title}{BERTweet-Hedge}.
\newblock \bibinfo{howpublished}{\url{https://huggingface.co/ChrisLiewJY/BERTweet-Hedge}}.
\newblock
\newblock
\shownote{Accessed: 2025-09-30}.


\bibitem[Liu et~al\mbox{.}(2019)]%
        {liu2019roberta}
\bibfield{author}{\bibinfo{person}{Yinhan Liu}, \bibinfo{person}{Myle Ott}, \bibinfo{person}{Naman Goyal}, \bibinfo{person}{Jingfei Du}, \bibinfo{person}{Mandar Joshi}, \bibinfo{person}{Danqi Chen}, \bibinfo{person}{Omer Levy}, \bibinfo{person}{Mike Lewis}, \bibinfo{person}{Luke Zettlemoyer}, {and} \bibinfo{person}{Veselin Stoyanov}.} \bibinfo{year}{2019}\natexlab{}.
\newblock \showarticletitle{Roberta: A robustly optimized bert pretraining approach}.
\newblock \bibinfo{journal}{\emph{arXiv preprint arXiv:1907.11692}} (\bibinfo{year}{2019}).
\newblock


\bibitem[Loper and Bird(2002)]%
        {loper2002nltk}
\bibfield{author}{\bibinfo{person}{Edward Loper} {and} \bibinfo{person}{Steven Bird}.} \bibinfo{year}{2002}\natexlab{}.
\newblock \showarticletitle{Nltk: The natural language toolkit}. In \bibinfo{booktitle}{\emph{Proceedings of the ACL-02 Workshop on Effective tools and methodologies for teaching natural language processing and computational linguistics}}. \bibinfo{pages}{63--70}.
\newblock


\bibitem[McInnes et~al\mbox{.}(2017)]%
        {mcinnes2017hdbscan}
\bibfield{author}{\bibinfo{person}{Leland McInnes}, \bibinfo{person}{John Healy}, \bibinfo{person}{Steve Astels}, {et~al\mbox{.}}} \bibinfo{year}{2017}\natexlab{}.
\newblock \showarticletitle{hdbscan: Hierarchical density based clustering.}
\newblock \bibinfo{journal}{\emph{J. Open Source Softw.}} \bibinfo{volume}{2}, \bibinfo{number}{11} (\bibinfo{year}{2017}), \bibinfo{pages}{205}.
\newblock


\bibitem[Mongeon et~al\mbox{.}(2023)]%
        {mongeon2023open}
\bibfield{author}{\bibinfo{person}{Philippe Mongeon}, \bibinfo{person}{Timothy~D Bowman}, {and} \bibinfo{person}{Rodrigo Costas}.} \bibinfo{year}{2023}\natexlab{}.
\newblock \showarticletitle{An open data set of scholars on Twitter}.
\newblock \bibinfo{journal}{\emph{Quantitative Science Studies}} \bibinfo{volume}{4}, \bibinfo{number}{2} (\bibinfo{year}{2023}), \bibinfo{pages}{314--324}.
\newblock


\bibitem[Pedregosa et~al\mbox{.}(2011)]%
        {scikit-learn}
\bibfield{author}{\bibinfo{person}{F. Pedregosa}, \bibinfo{person}{G. Varoquaux}, \bibinfo{person}{A. Gramfort}, \bibinfo{person}{V. Michel}, \bibinfo{person}{B. Thirion}, \bibinfo{person}{O. Grisel}, \bibinfo{person}{M. Blondel}, \bibinfo{person}{P. Prettenhofer}, \bibinfo{person}{R. Weiss}, \bibinfo{person}{V. Dubourg}, \bibinfo{person}{J. Vanderplas}, \bibinfo{person}{A. Passos}, \bibinfo{person}{D. Cournapeau}, \bibinfo{person}{M. Brucher}, \bibinfo{person}{M. Perrot}, {and} \bibinfo{person}{E. Duchesnay}.} \bibinfo{year}{2011}\natexlab{}.
\newblock \showarticletitle{Scikit-learn: Machine Learning in {P}ython}.
\newblock \bibinfo{journal}{\emph{Journal of Machine Learning Research}}  \bibinfo{volume}{12} (\bibinfo{year}{2011}), \bibinfo{pages}{2825--2830}.
\newblock


\bibitem[Priem et~al\mbox{.}(2022)]%
        {priem2022openalex}
\bibfield{author}{\bibinfo{person}{Jason Priem}, \bibinfo{person}{Heather Piwowar}, {and} \bibinfo{person}{Richard Orr}.} \bibinfo{year}{2022}\natexlab{}.
\newblock \showarticletitle{OpenAlex: A fully-open index of scholarly works, authors, venues, institutions, and concepts}.
\newblock \bibinfo{journal}{\emph{arXiv preprint arXiv:2205.01833}} (\bibinfo{year}{2022}).
\newblock


\bibitem[Rangel and Rosso(2019)]%
        {rangel2019overview}
\bibfield{author}{\bibinfo{person}{Francisco Rangel} {and} \bibinfo{person}{Paolo Rosso}.} \bibinfo{year}{2019}\natexlab{}.
\newblock \showarticletitle{Overview of the 7th author profiling task at PAN 2019: bots and gender profiling in twitter}.
\newblock \bibinfo{journal}{\emph{Working notes papers of the CLEF 2019 evaluation labs}}  \bibinfo{volume}{2380} (\bibinfo{year}{2019}), \bibinfo{pages}{1--7}.
\newblock


\bibitem[Reimers and Gurevych(2019)]%
        {reimers-2019-sentence-bert}
\bibfield{author}{\bibinfo{person}{Nils Reimers} {and} \bibinfo{person}{Iryna Gurevych}.} \bibinfo{year}{2019}\natexlab{}.
\newblock \showarticletitle{Sentence-BERT: Sentence Embeddings using Siamese BERT-Networks}. In \bibinfo{booktitle}{\emph{Proceedings of the 2019 Conference on Empirical Methods in Natural Language Processing}}. \bibinfo{publisher}{Association for Computational Linguistics}.
\newblock
\urldef\tempurl%
\url{https://arxiv.org/abs/1908.10084}
\showURL{%
\tempurl}


\bibitem[Rocca and Yarkoni(2022)]%
        {rocca2022language}
\bibfield{author}{\bibinfo{person}{Roberta Rocca} {and} \bibinfo{person}{Tal Yarkoni}.} \bibinfo{year}{2022}\natexlab{}.
\newblock \showarticletitle{Language as a fingerprint: Self-supervised learning of user encodings using transformers}. In \bibinfo{booktitle}{\emph{Findings of the Association for Computational Linguistics: EMNLP 2022}}. \bibinfo{pages}{1701--1714}.
\newblock


\bibitem[Schroff et~al\mbox{.}(2015)]%
        {schroff2015facenet}
\bibfield{author}{\bibinfo{person}{Florian Schroff}, \bibinfo{person}{Dmitry Kalenichenko}, {and} \bibinfo{person}{James Philbin}.} \bibinfo{year}{2015}\natexlab{}.
\newblock \showarticletitle{Facenet: A unified embedding for face recognition and clustering}. In \bibinfo{booktitle}{\emph{Proceedings of the IEEE conference on computer vision and pattern recognition}}. \bibinfo{pages}{815--823}.
\newblock


\bibitem[Shen(2021)]%
        {accuracybias}
\bibfield{author}{\bibinfo{person}{Lucas Shen}.} \bibinfo{year}{2021}\natexlab{}.
\newblock \bibinfo{title}{Measuring Political Media Slant Using Text Data}.
\newblock
\urldef\tempurl%
\url{https://www.lucasshen.com/research/media.pdf}
\showURL{%
\tempurl}


\bibitem[Shen(2022)]%
        {lex}
\bibfield{author}{\bibinfo{person}{Lucas Shen}.} \bibinfo{year}{2022}\natexlab{}.
\newblock \bibinfo{title}{{LexicalRichness: A small module to compute textual lexical richness}}.
\newblock
\href{https://doi.org/10.5281/zenodo.6607007}{doi:\nolinkurl{10.5281/zenodo.6607007}}


\bibitem[Soler-Company and Wanner(2017)]%
        {soler-company-wanner-2017-relevance}
\bibfield{author}{\bibinfo{person}{Juan Soler-Company} {and} \bibinfo{person}{Leo Wanner}.} \bibinfo{year}{2017}\natexlab{}.
\newblock \showarticletitle{On the Relevance of Syntactic and Discourse Features for Author Profiling and Identification}. In \bibinfo{booktitle}{\emph{Proceedings of the 15th Conference of the {E}uropean Chapter of the Association for Computational Linguistics: Volume 2, Short Papers}}, \bibfield{editor}{\bibinfo{person}{Mirella Lapata}, \bibinfo{person}{Phil Blunsom}, {and} \bibinfo{person}{Alexander Koller}} (Eds.). \bibinfo{publisher}{Association for Computational Linguistics}, \bibinfo{address}{Valencia, Spain}, \bibinfo{pages}{681--687}.
\newblock
\urldef\tempurl%
\url{https://aclanthology.org/E17-2108/}
\showURL{%
\tempurl}


\bibitem[van Schalkwyk and Dudek(2022)]%
        {van2022reporting}
\bibfield{author}{\bibinfo{person}{Fran{\c{c}}ois van Schalkwyk} {and} \bibinfo{person}{Jonathan Dudek}.} \bibinfo{year}{2022}\natexlab{}.
\newblock \showarticletitle{Reporting preprints in the media during the COVID-19 pandemic}.
\newblock \bibinfo{journal}{\emph{Public understanding of science}} \bibinfo{volume}{31}, \bibinfo{number}{5} (\bibinfo{year}{2022}), \bibinfo{pages}{608--616}.
\newblock


\bibitem[Vosoughi et~al\mbox{.}(2018)]%
        {vosoughi2018spread}
\bibfield{author}{\bibinfo{person}{Soroush Vosoughi}, \bibinfo{person}{Deb Roy}, {and} \bibinfo{person}{Sinan Aral}.} \bibinfo{year}{2018}\natexlab{}.
\newblock \showarticletitle{The spread of true and false news online}.
\newblock \bibinfo{journal}{\emph{science}} \bibinfo{volume}{359}, \bibinfo{number}{6380} (\bibinfo{year}{2018}), \bibinfo{pages}{1146--1151}.
\newblock


\bibitem[Walter et~al\mbox{.}(2019)]%
        {walter2019scientific}
\bibfield{author}{\bibinfo{person}{Stefanie Walter}, \bibinfo{person}{Ines L{\"o}rcher}, {and} \bibinfo{person}{Michael Br{\"u}ggemann}.} \bibinfo{year}{2019}\natexlab{}.
\newblock \showarticletitle{Scientific networks on Twitter: Analyzing scientists’ interactions in the climate change debate}.
\newblock \bibinfo{journal}{\emph{Public Understanding of Science}} \bibinfo{volume}{28}, \bibinfo{number}{6} (\bibinfo{year}{2019}), \bibinfo{pages}{696--712}.
\newblock


\bibitem[Wang et~al\mbox{.}(2020)]%
        {wang2020minilm}
\bibfield{author}{\bibinfo{person}{Wenhui Wang}, \bibinfo{person}{Furu Wei}, \bibinfo{person}{Li Dong}, \bibinfo{person}{Hangbo Bao}, \bibinfo{person}{Nan Yang}, {and} \bibinfo{person}{Ming Zhou}.} \bibinfo{year}{2020}\natexlab{}.
\newblock \showarticletitle{Minilm: Deep self-attention distillation for task-agnostic compression of pre-trained transformers}.
\newblock \bibinfo{journal}{\emph{Advances in neural information processing systems}}  \bibinfo{volume}{33} (\bibinfo{year}{2020}), \bibinfo{pages}{5776--5788}.
\newblock


\bibitem[Wei et~al\mbox{.}(2016)]%
        {wei2016learning}
\bibfield{author}{\bibinfo{person}{Wei Wei}, \bibinfo{person}{Gao Cong}, \bibinfo{person}{Chunyan Miao}, \bibinfo{person}{Feida Zhu}, {and} \bibinfo{person}{Guohui Li}.} \bibinfo{year}{2016}\natexlab{}.
\newblock \showarticletitle{Learning to find topic experts in Twitter via different relations}.
\newblock \bibinfo{journal}{\emph{IEEE Transactions on Knowledge and Data Engineering}} \bibinfo{volume}{28}, \bibinfo{number}{7} (\bibinfo{year}{2016}), \bibinfo{pages}{1764--1778}.
\newblock


\bibitem[Zangerle et~al\mbox{.}(2024)]%
        {zangerle2024overview}
\bibfield{author}{\bibinfo{person}{Eva Zangerle}, \bibinfo{person}{Maximilian Mayerl}, \bibinfo{person}{Martin Potthast}, {and} \bibinfo{person}{Benno Stein}.} \bibinfo{year}{2024}\natexlab{}.
\newblock \showarticletitle{Overview of the Multi-Author Writing Style Analysis Task at PAN 2024.}. In \bibinfo{booktitle}{\emph{CLEF (Working Notes)}}. \bibinfo{pages}{2424--2431}.
\newblock


\bibitem[Zeng et~al\mbox{.}(2019)]%
        {zeng2019detecting}
\bibfield{author}{\bibinfo{person}{Li Zeng}, \bibinfo{person}{Dharma Dailey}, \bibinfo{person}{Owla Mohamed}, \bibinfo{person}{Kate Starbird}, {and} \bibinfo{person}{Emma~S Spiro}.} \bibinfo{year}{2019}\natexlab{}.
\newblock \showarticletitle{Detecting journalism in the age of social media: three experiments in classifying journalists on twitter}. In \bibinfo{booktitle}{\emph{Proceedings of the International AAAI Conference on Web and Social Media}}, Vol.~\bibinfo{volume}{13}. \bibinfo{pages}{548--559}.
\newblock


\bibitem[Zhang et~al\mbox{.}(2025)]%
        {zhang2025online}
\bibfield{author}{\bibinfo{person}{Man Zhang}, \bibinfo{person}{Lisa Lundgren}, {and} \bibinfo{person}{Ha Nguyen}.} \bibinfo{year}{2025}\natexlab{}.
\newblock \showarticletitle{An Online Scientific Twitter World: Social Network Analysis of\# ScienceTwitter,\# SciComm, and\# AcademicTwitter}.
\newblock \bibinfo{journal}{\emph{Journalism and Media}} \bibinfo{volume}{6}, \bibinfo{number}{4} (\bibinfo{year}{2025}), \bibinfo{pages}{159}.
\newblock


\end{thebibliography}

\appendix

\section{Appendix}
\label{sec:app}
\subsection{Parameters}
\subsubsection{BertTopic}
For BertTopic we used the following configuration and hyperparameters: 'all-MiniLM-L6-v2' \citep{wang2020minilm} to embed the documents, UMap for dimensionality reduction with 50 neighbors, 2 components minimum distance of 0.0 and euclidian distance metric, HDBSCAN for clustering with a minimum cluster size of 5, euclidean distance metric, 'eom' for cluster selection method, and for vectorization TF-IDF was used with bm25 weighting and we reduced frequent words.
\begin{table*}[h!]
    \centering
    \scalebox{0.75}{
    \begin{tabular}{|c|c|c|c|}
    \hline
         Fold & Triplet Count Training & Triplet Count Validation & \makecell{Removed Users \\with duplicate texts}  \\\hline
         1 &  34,243 & 8,561 & 9 \\
         2 &  34,242 & 8,561 & 10 \\
         3 &  34,242 & 8,561 & 10 \\ % korrigiert
         4 &  34,243 & 8,561 & 9 \\
         5 &  34,247 & 8,562 & 4 \\
         6 &  34,242 & 8,561 & 10 \\
         7 &  34,243 & 8,561 & 9 \\
         8 &  34,241 & 8,561 & 11 \\
         9 &  34,243 & 8,561 & 9 \\
         10 &  34,242 & 8,561  & 10 \\ \hline
    \end{tabular}}
    \caption{Triplet summary over 10 folds for Tweet data. Each fold contains 41,653 instances in total. This fold is split into 80\% Train and 20\% Validation}
    \label{tab:triplet_stats_tweets}
\end{table*}

\begin{table*}[h!]
    \centering
    \scalebox{0.7}{
    \begin{tabular}{|c|c|c|c|c|c|}
    \hline
         Fold & Contrastive Pairs Training & Positive Pairs & Negative Pairs & Validation Pairs & \makecell{Removed Users with\\ duplicate texts}  \\\hline
         1 &  34,243 & 17,121 & 17,122 & 8,561 & 9 \\
         2 &  34,242 & 17,121 & 17,121 & 8,561 & 10 \\
         3 &  34,242 & 17,121 & 17,121 & 8,561 & 10 \\
         4 &  34,243 & 17,121 & 17,122 & 8,561 & 9 \\
         5 &  34,247 & 17,123 & 17,124 & 8,561 & 4 \\
         6 &  34,242 & 17,121 & 17,121 & 8,561 & 10 \\
         7 &  34,243 & 17,121 & 17,122 & 8,561 & 9 \\
         8 &  34,241 & 17,120 & 17,121 & 8,561 & 11 \\
         9 &  34,243 & 17,121 & 17,122 & 8,561 & 9 \\
         10 &  34,242 & 17,121 & 17,121 & 8,561 & 10 \\ \hline
    \end{tabular}}
    \caption{Contrastive summary over 10 folds for Tweet data. Each fold contains 41,653 instances in total. This fold is split into 80\% Train and 20\% Validation. The created dataset is balanced, }
    \label{tab:contrastive_stats_tweets}
\end{table*}

\subsubsection{Feature Based Models}
For each classifier, we used a random state of 42 and performed a hyperparameter search. 
For Random Forest we used bootstrap, max samples of 0.8, 300 estimators, a max depth of either None or 50, a minimum sample split of 2 or 5, minimum samples leaf of 1,2 or 4, maximum features of 0.5 or 'sqrt', criterion 'gini' and class weight of 'None' or 'balanced'.
For logistic regression we used 'lbgfgs', 'newton-cholesky' 'liblinear', 'newton-cg' or 'liblinear', penality of either 'elasticnet', l1 or l2 and C values of 0.001, 0.01, 0.1, 1 and 10.
For SVM, we used 'rbf', 'linear' or 'poly' kernel, a C value of 0.001, 0.1, 1 or 10, a gamma value of either 'scale', 0.01, 0.001, a degree of 2 or 3 and a 'coef0' of 0 or 1.
For AdaBoost, we used estimators of either 50,100,200 or 400, a learning rate of 0.01, 0.05, 0.1, 0.5 or 1.0, the estimator was a decision tree classifier with a maximum depth of 1,2 or 3 and minimum samples leaf of 1,2,4 or 10.

\subsection{Data creation}
The folds are not stratified in terms of positive and negative class as shown in Table \ref{tab:fold_statistics}. First, the authors in the K-Fold split were split into training and validation set. Authors which have the same representation (concatenated tweets or biographies) are not considered for triplet building in one fold. This deduplication is not considered across folds. The count of authors in the tweet datasets differs from the number of authors in the biography dataset because more there were more biographies which were not usable, so more authors had to be removed. 

\begin{table}[h]
    \centering
    \begin{tabular}{|c|c|c|}
    \hline
         Fold & Non-Scientists & Scientists  \\ \hline
         0 & 21,369 & 21,444 \\
         1 & 21,377 & 21,436 \\
         2 & 21,454 & 21,359 \\
         3 & 21,428 & 21,385 \\
         4 & 21,343 & 21,470 \\
         5 & 21,408 & 21,405 \\
         6 & 21,405 & 21,408 \\
         7 & 21,406 & 21,407 \\
         8 & 21,444 & 21,369 \\ 
         9 & 21,431 & 21,382 \\ \hline
    \end{tabular}
    \caption{Statistics for each fold. Folds will be split (not stratified) into training (80\%) and validation set (20\%).}
    \label{tab:fold_statistics}
\end{table}

\subsubsection{Triplet Creation}
The triplets were created by following the work of Rocca et al. \citep{rocca2022language}
We build one triplet per author and tracked duplicate positive pairs. Every Author was chosen as anchor. Based on the label of the anchor (0 for non-scientist, 1 for scientist) a positive sample (same label as anchor) was randomly selected and a negative sample (different label as anchor) was randomly selected. Constraint: Do not create triplets where the positive sample is the anchor paired with the same anchor as positive: $<a,p,n>$ != $<p,a,n>$. Choosing the negative component is random, this means one negative author can be used multiple or zero times. Statistics are shown in Table \ref{tab:triplet_stats_tweets} and \ref{tab:triplet_stats_bio}

\subsubsection{Creation of contrastive Pairs}
When building contrastive pairs, we build the same amount of positive and negative pairs. Furthermore, we build the same amount of instances of contrastive pairs as triplets.
To create positive and negative pairs, users were randomly paired. Each pair was stored so that no duplicate pairs were created. The amount of positive and negative pairs was set to the half amount of total authors so that the resulting dataset is balanced. One author was represented either by the concatenation of all of his tweets or biographical information.
No deduplication as in Triplet creation was performed, but the contrastive pairs (e.g. p1 and p2) are logged, so that no p2 and p1 is created. Statistics for Contrastive data for Tweet and Biography data are shown in Tables \ref{tab:contrastive_stats_tweets} and \ref{tab:contrastive_stats_bio}

\begin{table*}[h!]
    \centering
    \scalebox{0.8}{
    \begin{tabular}{|c|c|c|c|}
    \hline
         Fold & Triplet Count Training & Triplet Count Validation &  \makecell{Removed Users with\\ duplicate texts}  \\\hline
         1 &  30,891 & 7,723 & 67 \\
         2 &  30,907 & 7,727 & 69 \\
         3 &  30,869 & 7,718 & 62 \\ % korrigiert
         4 &  30,871 & 7,718 & 69 \\
         5 &  30,892 & 7,723 & 68 \\
         6 &  30,857 & 7,715 & 76 \\
         7 &  30,871 & 7,718 & 66 \\
         8 &  30,872 & 7,718 & 69 \\
         9 &  30,854 & 7,714 & 63 \\
         10 &  30,879 & 7,720  & 66 \\ \hline
    \end{tabular}}
    \caption{Triplet summary over 10 folds for \textbf{biography}. Each fold contains 41,653 instances in total. This fold is split into 80\% Train and 20\% Validation}
    \label{tab:triplet_stats_bio}
\end{table*}

\begin{table*}[h!]
    \centering
    \scalebox{0.8}{
    \begin{tabular}{|c|c|c|c|c|c|}
    \hline
         Fold & Contrastive Pairs Training & Positive Pairs & Negative Pairs & Validation Pairs & \makecell{Removed Users with\\ duplicate texts}  \\\hline
         1 &  30,891 & 15,445 & 15,446 & 7,723 & 67 \\
         2 &  30,907 & 15,453 & 15,454 & 7,727 & 69 \\
         3 &  30,869 & 15,434 & 15,435 & 7,718 & 62 \\
         4 &  30,871 & 15,435 & 15,436 & 7,718 & 69 \\
         5 &  30,892 & 15,446 & 15,446 & 7,723 & 68 \\
         6 &  30,857 & 15,428 & 15,429 & 7,715 & 76 \\
         7 &  30,871 & 15,435 & 15,436 & 7,718 & 66 \\
         8 &  30,872 & 15,436 & 15,436 & 7,718 & 69 \\
         9 &  30,854 & 15,427 & 15,427 & 7,714 & 63 \\
         10 &  30,879 & 15,439 & 15,440 & 7,720 & 66 \\ \hline
    \end{tabular}}
    \caption{Contrastive summary over 10 folds for \textbf{biography data}. Each fold contains 41,653 instances in total. This fold is split into 80\% Train and 20\% Validation.}
    \label{tab:contrastive_stats_bio}
\end{table*}

\end{document}